\documentclass[preprint,12pt]{elsarticle}
\usepackage{booktabs}  

\usepackage{subcaption}
\usepackage{amsmath,amssymb,amsfonts}
\usepackage{algorithmic,algorithm}
\usepackage{graphicx}
\usepackage{hyperref} 
\usepackage{siunitx}

\hypersetup{colorlinks=true,
            linkcolor = blue,
            urlcolor  = blue,
            citecolor = blue,
            anchorcolor = blue}

\begin{document}

\begin{frontmatter}

\title{Gaussian Derivative Change-point Detection for Early Warnings of Industrial System Failures}

\author{Hao Zhao}
\author{Rong Pan\corref{cor1}}
\ead{Rong.Pan@asu.edu}
\cortext[cor1]{Corresponding author}

\affiliation{organization={School of Computing and Augmented Intelligence, Arizona State University},
            addressline={699 S Mill Ave}, 
            city={Tempe},
            postcode={85281}, 
            state={AZ},
            country={USA}}

\begin{abstract}
An early warning of future system failure is essential for conducting predictive maintenance and enhancing system availability. This paper introduces a three-step framework for assessing system health to predict imminent system breakdowns.  First, the Gaussian Derivative Change-Point Detection (GDCPD) algorithm is proposed for detecting changes in the high-dimensional feature space. GDCPD conducts a multivariate Change-Point Detection (CPD) by implementing Gaussian derivative processes for identifying change locations on critical system features, as these changes eventually will lead to system failure. To assess the significance of these changes, Weighted Mahalanobis Distance (WMD) is applied in both offline and online analyses. In the offline setting, WMD helps establish a threshold that determines significant system variations, while in the online setting, it facilitates real-time monitoring, issuing alarms for potential future system breakdowns. Utilizing the insights gained from the GDCPD and monitoring scheme, Long Short-Term Memory (LSTM) network is then employed to estimate the Remaining Useful Life (RUL) of the system. The experimental study of a real-world system demonstrates the effectiveness of the proposed methodology in accurately forecasting system failures well before they occur. By integrating CPD with real-time monitoring and RUL prediction, this methodology significantly advances system health monitoring and early warning capabilities.

\end{abstract}





\end{frontmatter}


\section{Introduction}
\label{sec:introduction}
Early warnings of industrial equipment breakdowns are essential to predictive maintenance and system availability. The main idea behind the efficacy of an early warning system is the statistical Change-Point Detection (CPD), which is an analytical framework that identifies abrupt changes in the industrial process. Accurately locating these changes is crucial for understanding and addressing the underlying process dynamics, which can also significantly impact any machine learning algorithm's predictive performance. CPD plays a pivotal role in the early warning system; however, it faces challenges in analyzing complex, high-dimensional data, which is typical in modern industrial systems. These challenges include determining when and where changes occur and identifying the specific features that are sensitive to process changes \cite{pan2012bayesian,steward2016bayesian}.

Most existing CPD techniques often depend on predefined parametric model assumptions about how the data was generated, which include the data distribution model and the process change characteristics, such as mean shift or trend detection. Examples of such approaches include the Generalized Likelihood Ratio (GLR) test \cite{dette2020likelihood}, Cumulative Sum (CUSUM) test \cite{lee2020hybrid}, and Bayesian change-point analysis \cite{wu2024unsupervised}. The efficacy of these parametric approaches heavily depends on the validity of their assumptions. If the chosen model does not capture the actual underlying change mechanism, the CPD accuracy will be undermined. In contrast, non-parametric approaches, which rely on data-driven Machine Learning (ML) methods and declare change-points when a statistic exceeds a certain threshold, offer an alternative approach that can avoid mismatches between model assumptions and real data more easily. One common choice is Maximum Mean Discrepancy (MMD), which measures the distance between two distributions using test functions in a Reproducing Kernel Hilbert Space (RKHS) \cite{sinn2012detecting}. But when applying kernel MMD to identify mean shifts in Gaussian distributions, the test power reduces polynomially with increasing data dimension. To address this problem, Wang \emph{et al.} \cite{wang2021two} employed dimension reduction techniques, using the projected Wasserstein distance to enhance the detection method in high-dimensional scenarios. Xie \emph{et al.} \cite{xie2021sequential} proposed a sequential change detection test based on a weighted $\ell_{2}$ divergence, focusing on empirical distribution distance from four consecutive data windows around the change-point. Other test statistics, like Sinkhorn divergence \cite{wang2022data}, Mahalanobis distance (MD) \cite{xie2012change}, f-divergence \cite{dwivedi2022discriminant}, and classifier-based distances \cite{londschien2023random} also offer different alternatives for statistical testing in CPD within high-dimensional settings. For a more extensive discussion on CPD methodologies, see \cite{truong2020selective, van2020evaluation}.

As a non-parametric model, Gaussian Processes (GPs) demonstrate enhanced robustness and flexibility across different data types and scenarios involving change-points \cite{saatcci2010gaussian}. Keshavarz \emph{et al.} \cite{keshavarz2018optimal} introduced a novel detection method based on the Generalized Likelihood Ratio Test (GLRT), which achieves near-optimal performance by exploiting the data covariance structure in GP. Caldarelli \emph{et al.} \cite{caldarelli2022adaptive} presented the algorithm that dynamically adjusts GP hyperparameters to better adapt to data shifts, significantly improving CPD in complex datasets. Zhao and Pan \cite{zhao2023active} used the GP-based active learning method to guide data queries around likely locations of change-points, addressing the issue of costly and time-intensive data collection for CPD.  

Typically, CPD can be classified into online and offline approaches. Offline methods analyze the complete dataset retrospectively to identify change-points, aiming to detect all changes in batch mode. Conversely, online methods are designed for real-time analysis by evaluating data points sequentially to promptly detect changes. As our goal is to establish a comprehensive early warning system, it is crucial to implement an online monitoring mechanism for immediate change detection as well as accurately predict the Remaining Useful Life (RUL) upon triggering a system warning. This approach enhances system health awareness, allowing for effective maintenance planning.

RUL prediction focuses on estimating the time until a system or component fails. Most of the research in RUL prognostics employ either statistical model-based or ML-based techniques \cite{wang2020remaining}. Long Short-Term Memory (LSTM) networks are a group of deep neural network-based ML methods, which are gaining popularity in recent literature given their ability to sequentially model and manage long-term dependencies using a unique gating mechanism. In addition, several studies have proposed methods to handle differences in data distributions between training and testing environments, which often arise in real-world applications due to varying operating condition. For instance, domain adaptation techniques are used in LSTM to make models generalizable across different machines or conditions by learning domain-invariant features \cite{da2020remaining, fu2021deep}. Other studies combine LSTM with neural network layers, such as Time Delay Neural Networks (TDNN) and convolutional layers, to enhance RUL prediction accuracy by capturing short-term dependencies and specific degradation trends \cite{zhang2023two, dong2023deep}. Attention mechanisms are also integrated into LSTM-based models to improve performance by focusing on critical features and time steps \cite{liu2022aircraft, zhu2024adaptive}. These methods are helpful when dealing with complex sensor data, where not all features contribute equally to the degradation process. In lithium-ion battery applications, some methods are designed to reduce computational complexity while maintaining prediction quality. Wang \emph{et al.} \cite{wang2023flexible} and Lyu \emph{et al.} \cite{lyu2023parallel} introduced techniques like poly-cell structures or state fusion to focus on the most important data while reducing the computational load. Furthermore, other approaches focus on detecting the start of degradation or generating structured health representations to enhance prediction accuracy, particularly in systems with non-linear degradation processes \cite{shi2021dual, zhu2024contrastive}.


Traditional RUL prediction models are designed for gradual system degradations and they are not suitable for predicting sudden system failures. In contrast, our model is specifically designed to detect process changes before a sudden system failure occurs. By statistically identifying these changes and treating them as the indication of upcoming system failures, we provide an approach to address a problem that conventional methods often overlook.

Building on the strengths of GP in performing multivariate CDP tasks and leveraging LSTM for RUL prediction, we have developed an advanced early warning system for monitoring and predicting the health of a sensor-embedded manufacturing process. This system is built upon the detection of change-points using a multivariate GP with derivative information in an offline setting, which we refer to as Gaussian Derivative CPD (GDCPD). GDCPD utilizes the Gaussian derivative process to detect sudden shifts in processes, and it also uses Automatic Relevance Determination (ARD) as a feature selection tool, where a smaller length-scale value in ARD indicates the feature being more influential. Thus, GDCPD can be viewed as a target-focused approach to establishing a system health monitoring scheme. 

\begin{figure}[t!]
\centerline{\includegraphics[width=1.0\textwidth]{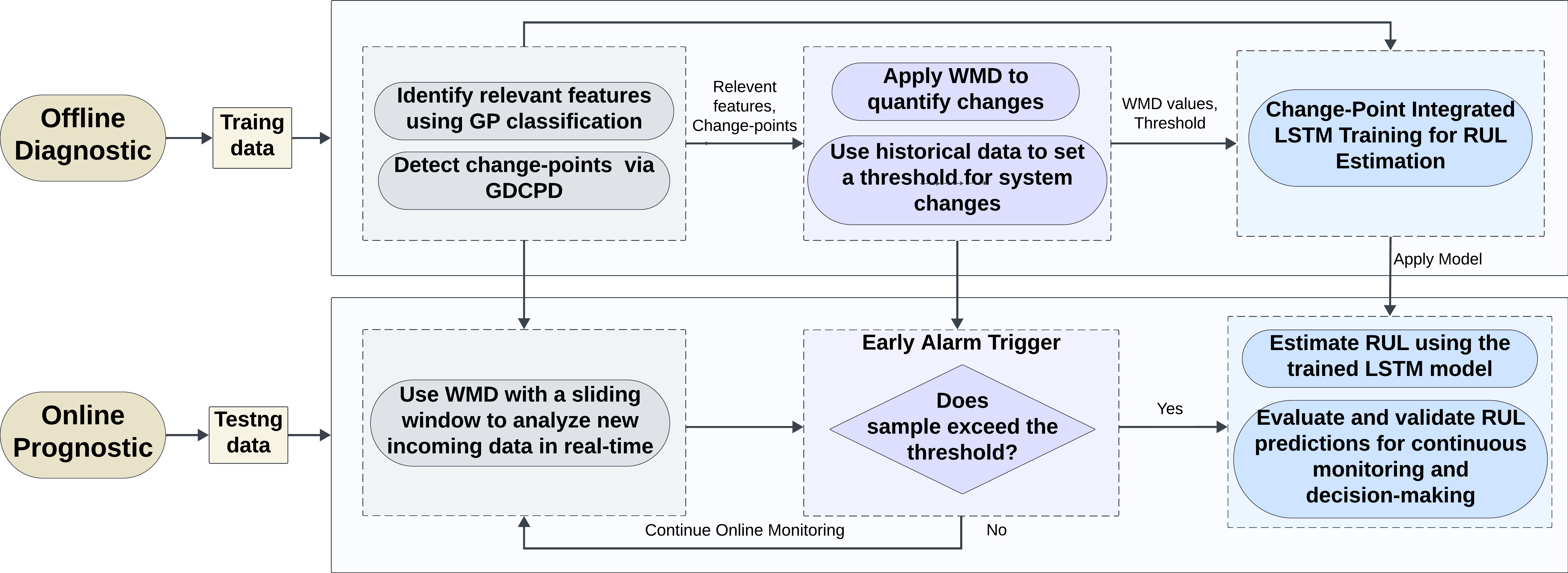}}
\caption{The flowchart of the proposed early warning system for monitoring and predicting system failures. The system health and RUL are analyzed through two phases: the offline diagnostic phase and the online prognostic phase. In the offline phase, training data is processed via GDCPD to identify important features and change points on these features. They are used to set the threshold for monitoring statistics and to build LSTM predictive models. For online prognostics, a sliding window mechanism is used to detect early alarms, along with the trained LSTM models for RUL prediction. This structured approach ensures dynamic, real-time system health monitoring and failure prediction.}
\label{fig:flowchart}
\end{figure}

With a robust CPD method established, we can set the early warning threshold using Weighted Mahalanobis Distance (WMD). This threshold is a critical marker for indicating when the system state deviates beyond its normal range. Finally, when the early warning system is operated online, a sequential process monitoring for detecting process abrupt changes is implemented. We incorporate LSTM networks for the RUL estimation when a warning alarm is issued. Figure~\ref{fig:flowchart} depicts the complete early warning system.

In this paper, we use the process monitoring data from a paper manufacturing industry to demonstrate our methodology. Figure~\ref{fig:channel} plots the signal values from one of 61 sensors deployed on the production line. One may notice that some signal shifts between the two manufacturing breakdown events; however, these shifts are irregular and noisy, thus challenging for system failure warning. We aim to utilize all sensor channels and provide a comprehensive approach that integrates multivariate CPD, WMD quantification, and LSTM for predictive modeling to create an advanced system health monitoring and early failure warning system. The key contributions of this work include:

\begin{figure}[t!]
\centerline{\includegraphics[width=0.69\textwidth]{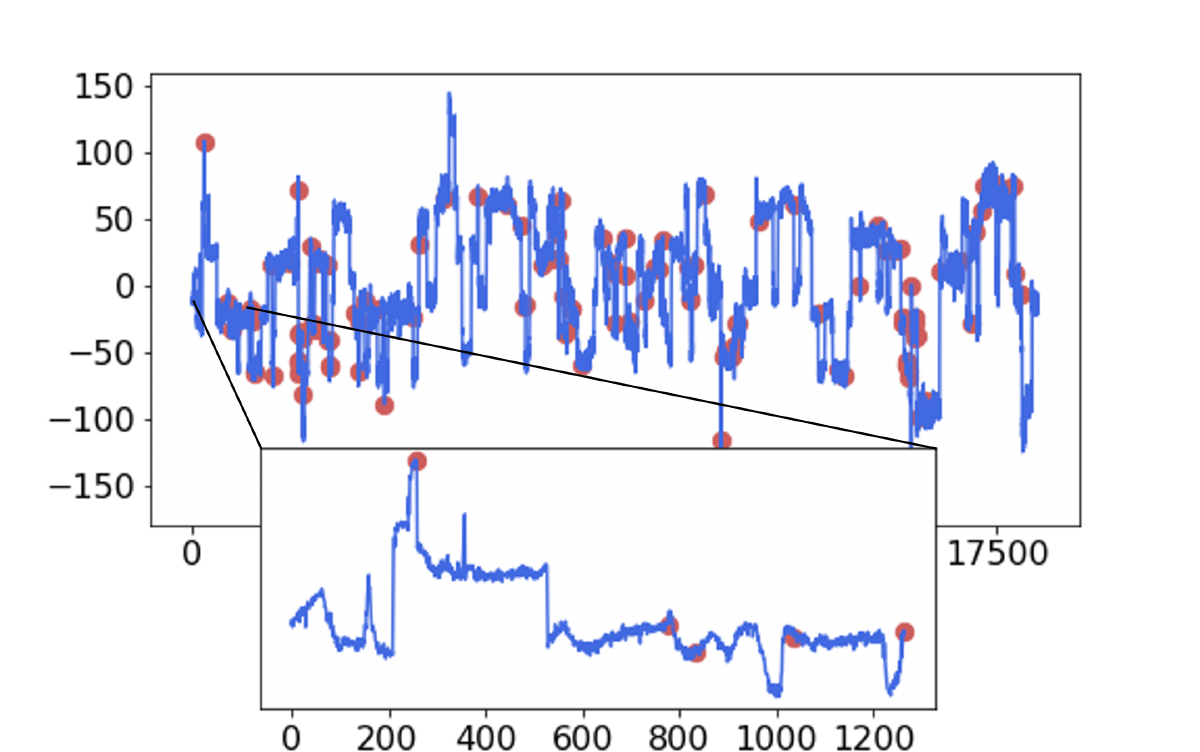}}
\caption{Sensor Signals of Paper Manufacturing Machine. The top figure illustrates the signal time series from one of  61 sensors, with red dots marking the times of system breakdowns. The figure below zooms in the first five breakdowns. Note that after each breakdown the system will be repaired and restarted. The repair periods are not included in the figure.}
\label{fig:channel}
\end{figure}

\begin{enumerate}[1.]
    \item We developed GDCPD, a novel data-driven nonparametric change-point detection approach, which is designed for handling high-dimensional data;
    \item We established a system monitoring threshold based on the temporal features learned via GDCPD. Our approach is highly explainable and provides clear insights into which features and when these features may cause future failures; 
    \item We improved RUL prediction by integrating LSTM into the early warning system. By leveraging the data adjusted through early warnings, we enhanced the accuracy and reliability of RUL estimations, enabling timely preventive actions and better maintenance planning.
\end{enumerate}

The structure of this paper is as follows: We begin with a review of related work in Section \ref{RelatedWork}. Section \ref{MCPD} introduces the GDCPD model. In Section \ref{EW}, we detail the setup of our early warning system. Section \ref{experiment} presents an experimental study to demonstrate the effectiveness of our methodology. Finally, the conclusions are drawn in Section \ref{Conclusion}.

\section{Related Work}
\label{RelatedWork}

In the rich field of CPD and early warning systems, several foundational methods and concepts are relevant to our method. This section aims to elucidate the relationship between our methodology and past research work.

Considering the GP with derivatives in our model, we were inspired by the idea of Knowledge-Gradient (KG) in Bayesian optimization. The KG approach \cite{wu2017bayesian} involves using gradient data from GPs to speed up the optimization process. The gradient can provide more information about how the system changed, and thus plays a critical role in modeling dynamic systems \cite{padidar2021scaling}. In these studies, the function values and gradient vectors are assumed to be directly observed. In contrast, our setup does not rely on direct observation of the gradients, and the Gaussian derivative process is derived by differentiating the kernel function.

The CPD solution for online process monitoring can be viewed as a diagnostic tool, as potential changes or anomalies in the process can be promptly identified when significant deviations from the normal process state are detected. Multivariate statistical process control (MSPC) has been applied to process monitoring in various industries, including the chemical industry, pharmaceutical industry, metal fabrication, paper production, and food production \cite{ramos2021multivariate}. Techniques such as Principal Component Analysis (PCA), Canonical Correlation Analysis (CCA), and Independent Component Analysis (ICA) have played a key role in MSPC. PCA allows reliable feature extraction from datasets by projecting data onto a lower-dimensional sub-space, which simplifies analysis and change detection \cite{chaleshtori2024novel, de2021framework}. ICA offers a solution when PCA is not ideally suited, aiming to identify independent signals that can reconstruct complex datasets \cite{tharwat2021independent}. CCA and Multivariate Projection to Latent Structures (MPLS) enhance change detection by focusing on maximizing data correlations and covariance between data arrays \cite{gloaguen2022multiway}. Besides that, many machine learning techniques, e.g., Bayesian methods \cite{wu2024unsupervised}, Artificial Neural Networks (ANN) \cite{zhao2024online}, and Hidden Markov Model (HMM) \cite{soleimani2021integration}, have also been investigated for process change detection and diagnosis.

There have been a few developing research efforts recognizing the need for CPD methods to improve RUL estimation. For instance, Shi \emph{et al.} \cite{shi2021dual} introduced a dual-LSTM framework that combines CPD with RUL prediction. In their approach, cross-entropy is utilized to identify changes in the first LSTM model, which are then used to predict RUL in the second LSTM model. However, their method treats change-points just as split points for distinguishing between normal and degraded machine health conditions, and mainly focuses on late-stage RUL estimation. Similarly, Arunan \emph{et al.} \cite{arunan2024change} proposed a CPD-integrated RUL estimation model under variable operating conditions, but their detection algorithm utilized a parametric method. It relies on specific assumptions about the data distribution and involves estimating parameters from the data to perform the analysis. 

In contrast, our proposed GDCPD method is a nonparametric algorithm, and it is adaptable to a wide range of data types and change characteristics. This flexibility allows it to handle non-stationary and irregular data patterns commonly found in industrial applications. Notably, GDCPD can leverage the ARD framework to extract temporal variations related to failures across multivariate signals. Besides, our work provides a comprehensive framework for building an online early warning system. During offline model development, we dynamically construct monitoring statistics and early warning thresholds based on the temporal features learned via GDCPD. The detected change-points then trigger early alarms for the training of an LSTM-based RUL estimation model. During online monitoring, the temporal dynamics of industrial signals are continuously monitored for any breach of the thresholds established in offline training. Upon detecting an early alarm, the system RUL is estimated using the pre-trained offline model for timely preventive actions.


\section{Multivariate Change-point Detection}
\label{MCPD}


\subsection{Problem Formulation}
The offline multivariate CPD algorithm is designed to identify change locations when a system experiences significant changes across multiple variables or dimensions simultaneously. Suppose there are $N$ observations $\mathcal{D}=\left\{\boldsymbol{x}_i, y_i\right\}_{i=1}^N$, where $\boldsymbol{x}_i\in \mathbb{R}^d$ is a vector of input from $d$ different system variables or features, and $y_i \in \mathbb{R}$ is the output (i.e., system responses) follow an unknown function $f(y_i|\boldsymbol{x}_i)$. The goal is to detect change-points $\mathcal{T}=\{\tau_1, \tau_2,\cdots,\tau_K\}$, $\tau_i<\tau_j$ for $i<j$, and the integer $K$ corresponds to the total number of changes. The data sequence is thus partitioned into $K+1$ segments, each reflecting the hidden system states by the joint statistical behavior of these variables.

\subsection{Gaussian Process Classification}
GP is a non-parametric model that captures inherent stochasticity in system performance and allows for quantification of uncertainties associated with the state of the system. The system's output distribution $\boldsymbol{f}$ can be described through a mean function $\mu: \mathbb{R}^d \rightarrow \mathbb{R}$ and a covariance function $k: \mathbb{R}^d \times \mathbb{R}^d \rightarrow \mathbb{R}$, expressed as:
\begin{equation}
\label{eq:1}
\boldsymbol{f}\sim \mathcal{MGP}(\mu(\cdot), k(\cdot, \cdot)).
\end{equation}

In GP Classification, we consider a set of $N$ training inputs $\boldsymbol{X}=[\boldsymbol{x}_1, \ldots, \boldsymbol{x}_N]^\top$ and their corresponding class labels $\boldsymbol{y}=[y_1, \ldots, y_N]^\top$, where $y_i \in\{0,1\}$ for binary classification. The goal is to predict the class membership probability for a new test point $\boldsymbol{x}^\prime$. We introduce a latent function $\boldsymbol{f}(\boldsymbol{x})$ and model the probability of observed labels using a link function. For binary classification, we use the probit function $\Phi$, so the class membership probability is given by:
\begin{equation} P(y = 1 | \boldsymbol{x}) = \Phi(f(\boldsymbol{x})). \end{equation}

Given the training data $\mathcal{D}=\left\{\boldsymbol{x}_i, y_i\right\}_{i=1}^N$, the prior over the latent function values $\boldsymbol{f}$ at the training inputs is $\boldsymbol{f} \sim \mathcal{N}(\boldsymbol{\mu},\boldsymbol{K})$. Then, for a new test input $\boldsymbol{x}^{\prime}$, the predictive distribution of the latent function $f^{\prime}$ is given by:
\begin{equation} p(f^{\prime} | \boldsymbol{x}^{\prime}, \boldsymbol{X}, \boldsymbol{y}) = \int p(f^{\prime} | \boldsymbol{x}^{\prime}, \boldsymbol{X}, \boldsymbol{f})  p(\boldsymbol{f} | \boldsymbol{X}, \boldsymbol{y}) d\boldsymbol{f}. \end{equation}

Thus, the predictive probability for the class label is:
\begin{equation} P(y^{\prime} = 1 | \boldsymbol{x}^{\prime}, \boldsymbol{X}, \boldsymbol{y}) = \int \Phi(f^{\prime})  p(f^{\prime} | \boldsymbol{x}^{\prime}, \boldsymbol{X}, \boldsymbol{y})  df^{\prime}. \end{equation}

Because the likelihood $p(y|\boldsymbol{f})$ is non-Gaussian due to the discrete nature of class labels, the posterior distribution $p(\boldsymbol{f}|y,x)$ cannot be computed analytically. To address this issue, we can use approximation methods such as the Laplace approximation, expectation propagation \citep{kuss2005assessing}, or variational inference \citep{hensman2015scalable} to approximate the posterior.

To further diagnose multivariate inputs, we employ ARD to identify the most relevant features in the input space. ARD adjusts the length-scales $\lambda_d$ of different variables in the covariance function, effectively determining which variables significantly influence the system's output. The ARD kernel is expressed as:
\begin{equation}
k_{\mathrm{ARD}}(x, x^\prime)=v^2 \exp \left(-\frac{1}{2} \sum_{d=1}^D\left(\frac{x_d-x^{\prime}_d}{\lambda_d}\right)^2\right),
\end{equation}
where the hyperparameter $v^2$ controls the overall variability, and the individual length-scale parameters $\lambda_d$ allow the functions to vary at different scales across different variables.
This mechanism allows for a more focused analysis by emphasizing the impact of important features while diminishing the influence of less relevant ones. Through ARD, the multi-input GP classification model becomes not only more interpretable but also more efficient, as it avoids overfitting by disregarding unnecessary features.

\subsection{Gaussian Process Derivatives }

We can also utilize GP Regression to investigate the dynamics of these feature processes. Given that mathematical differentiation is a linear operator, the derivative of a GP retains the characteristics of GP \cite{rasmussen2006gaussian}. Consider a GP over a function $\boldsymbol{g}(t)$, where $t$ is the timestamp of feature signals. The function $\boldsymbol{g}(t)$ is modeled as $\boldsymbol{g}(t) \sim \mathcal{GP}(\mu(t), k(t, t'))$. In the context of regression, we have observations $\mathcal{D}=\left\{t_i, \boldsymbol{z}_i\right\}_{i=1}^N$, where $\boldsymbol{z}_i=\boldsymbol{g}(t_i)+\boldsymbol{\epsilon}_i$ and $\boldsymbol{\epsilon}_i$ are independent Gaussian noise terms with variance $\sigma^2$. The joint distribution of the observed outputs $\boldsymbol{z}$ and the derivative values
$\boldsymbol{z}^{\nabla}(t)$ at test points $t^{\prime}$ is:

\begin{equation}\begin{aligned}
\boldsymbol\mu^{\nabla}(t)&=\left[\begin{array}{c}
\boldsymbol{\mu}(t) \\
\partial_{t} \boldsymbol{\mu}(t)
\end{array}\right], \\
\boldsymbol{k}^{\nabla}\left(t, t^{\prime}\right)&=\left[\begin{array}{cc}
\boldsymbol{k}\left(t, t\right) & \left(\partial_{t^{\prime}} \boldsymbol{k}\left(t, t^{\prime}\right)\right)^T \\
\partial_{t}\boldsymbol{k}\left(t^{\prime}, t\right) & \partial^2 \boldsymbol{k}\left(t^{\prime}, t^{\prime}\right)
\end{array}\right],
\end{aligned}\end{equation}
where $\boldsymbol{k}(t, t)$ is an $n \times n$ covariance matrix whose entries contain all pairwise covariances between training points, $\boldsymbol{k}(t, t^{\prime}) = [k(t^{\prime}, t_1), \ldots, k(t^{\prime}, t_n)]^{\top} \in \mathbb{R}^n$ represents the covariances between the test point $t^{\prime}$ and all training points. $\boldsymbol{k}(t^{\prime}, t^{\prime})$ denotes the self-similarity matrix for the test samples. $\partial_{t^{\prime}} \boldsymbol{k}(t, t^{\prime}) := \left[\frac{\partial k(t, t_1)}{\partial t^{\prime}}, \ldots, \frac{\partial k(t, t_n)}{\partial t^{\prime}}\right]$ is the vector of first-order partial derivatives, and $\partial^2 k(t^{\prime}, t^{\prime}) := \left[\frac{\partial^2 k(t^{\prime}, t^{\prime})}{\partial t_i \partial t_j}, \ldots, \frac{\partial^2 k(t^{\prime}, t^{\prime})}{\partial t_i \partial t_j}\right]$ is the matrix of second-order mixed partial derivatives. Given the training data, the derivative of $\boldsymbol{g}$ follows a posterior distribution that is also a GP, with the mean and covariance determined as follows (see \ref{A} for the derivation):

\begin{equation}\begin{aligned}
   \boldsymbol\mu^\nabla(t^\prime) = &\partial_{t}\boldsymbol{k}\left(t, t^{\prime}\right)\boldsymbol{C}^{-1} \boldsymbol{z}, \\
   \boldsymbol{k}^\nabla\left(t, t^\prime\right)  = &\partial^2 \boldsymbol{k}\left(t^{\prime}, t^{\prime}\right)-
\left(\partial_{t^{\prime}} \boldsymbol{k}\left(t, t^{\prime}\right)\right)^T\boldsymbol{C}^{-1} \partial_{t}\boldsymbol{k}\left(t, t^{\prime}\right),
\end{aligned}\end{equation}
where $\boldsymbol{C}=(\boldsymbol{k}(t, t)+\sigma^2\boldsymbol{I})$, and  $\boldsymbol{I}$ denotes an identity matrix.

The computing cost of GP derivatives is significant due to their cubic scaling with the number of observations $N$ and dimension $D$, denoted as $O(N^3D^3)$. However, this issue can be alleviated by sparse GP and other methods \cite{hensman2018variational, snelson2005sparse, wilson2015kernel}.

In this paper, we use the Radial Basis Function (RBF) kernel, which is defined by
\begin{equation}
k_{\mathrm{RBF}}(t, t^\prime)=\sigma^2 \exp \left(-\frac{\|t-t^\prime\|^2}{2 \ell^2}\right).
\end{equation}
The first-order partial derivatives and mixed partial derivatives of the RBF kernel are calculated as follows \cite{johnson2020kernel}:
\begin{equation}\begin{aligned} 
\frac{\partial k_{\mathrm{RBF}}\left(t, t^\prime\right)}{\partial t_i^{\prime}}=&\frac{t_i-t_i^{\prime}}{\ell^2} k_{\mathrm{RBF}}\left(t, t^{\prime}\right), \\ 
\frac{\partial^2 k_{\mathrm{RBF}}\left(t, t^{\prime}\right)}{\partial t_i \partial t_j^{\prime}}=&\frac{1}{\ell^4}\left(\ell^2 \delta_{i j}-\left(t_i-t_i^{\prime}\right)\left(t_j-t_j^{\prime}\right)\right) 
\times k_{\mathrm{RBF}}\left(t, t^{\prime}\right).
\end{aligned}\end{equation}
For the details of these derivations, see \ref{B}.

\subsection{Gaussian Derivative Change-point Detection}

\begin{algorithm}[t]
\caption{GDCPD}
\label{alg:GDCPD}
\renewcommand{\algorithmicrequire}{\textbf{Input:}}
\renewcommand{\algorithmicensure}{\textbf{Output:}}

\begin{algorithmic}[1]
    \REQUIRE      \hspace*{\algorithmicindent}{}\\
     Observation set: $\mathcal{D}$\\ 
            Number of change-points: $K$\\
    \ENSURE Change-points set: $\hat{\mathcal{T}}$
    \STATE Determine the relative dimensions via GP classification
    \STATE Fit GP derivative kernel with selected features
    \STATE Calculate $S_i$ based on Eq.\eqref{eq:7}
    \FOR{$i = 1, \cdots, K$}
        \STATE Determine $\hat{\tau}_k= \arg\max_{k \in [A, n-A]} |D(k, A)|$ for all potential change-points $\tilde{\tau}_k$
        \STATE Set $D(k, A) = 0$ for all $k \in  (\hat{\tau}_k - A, \hat{\tau}_k + A)$
        \STATE Add $\hat{\tau}_k$ to set of change-point $\hat{\mathcal{T}}$
    \ENDFOR
    \RETURN $\hat{\mathcal{T}}$
\end{algorithmic}
\end{algorithm}

GP serves as our base CPD model primarily due to its flexibility and non-parametric nature, which acts like a universal approximator, making it suitable for modeling a wide range of data types, including those with complicated or non-stationary features.

Building on the GP and ARD framework, we introduce the GDCPD method to address the challenges of multivariate CPD problems in the offline setting with knowing the number of change points, $K$ (see Algorithm \ref{alg:GDCPD}). Intuitively, when a continuous process has an abrupt change, its first-order derivative will increase to a maximal value at the change location, while the second-order derivative will reveal the function's curvature or concavity. Along with this reasoning, the derivative process of GP can be utilized to locate the changes in system behavior. By fitting a derivative kernel to the data, GDCPD captures variations in the system's output with respect to its inputs. To systematically identify all possible change points $\tilde{\tau}_k$, the GDCPD method iterates through the data, searching for the highest summed absolute derivatives $S_i$ across all relevant dimensions,
\begin{equation}
\label{eq:7}
     S_i = \sum_{d=1}^{D} |\nabla\boldsymbol{\mu}_{i,d}|, \quad \text{for all }\forall i.
\end{equation}
\begin{equation}
\label{eq:8}
    \tilde{\tau}_k = \arg\max_{i \notin \tilde{\mathcal{T}}} S_i.
\end{equation}
To ensure the identified change-points accurately reflect actual system changes, instead of artifacts caused by noise in the data, we employed a fixed window mean difference calculation around each potential change-point \cite{bertrand2011off}. This method helps confirm the significance of each potential change-point. The procedure involves calculating the difference between two empirical means, defined as:
\begin{equation}
\label{eq:9}
D(k, A)=\hat{\theta}(k, A)-\hat{\theta}(k-A, A),
\end{equation}
where $\hat{\theta}(k, A)$ is the empirical mean of $\boldsymbol{x}$ calculated within the sliding window $[k + 1, k + A]$. The difference $D(k, A)$ is associated with a sequence, denoted as $\{D(k, A)\}_{A\leq k\leq n-A}$, and $A$ is the window size.

The actual change-point, $\hat{\tau}_k$, is then determined by identifying the position where the absolute value of $D(k,A)$ is maximized:
\begin{equation}
\label{eq:10}
    \hat{\tau}_k =\arg \max _{k \in[A, n-A]}|D(k, A)|.
\end{equation}
Once a change point is detected, the surrounding area (defined by the window size $A$) is suppressed by setting differences to zero. This step prevents the same points from causing multiple detections and allows the algorithm to focus on finding the next significant change. The procedure continues until the set $\hat{\mathcal{T}}=\{\boldsymbol{x}_{\hat{\tau}_1}, \boldsymbol{x}_{\hat{\tau}_2}, \cdots,\boldsymbol{x}_{\hat{\tau}_K} \}$ is completed. 

Optimal kernel hyperparameters are determined by minimizing the negative log marginal likelihood with the L-BFGS optimizer. L-BFGS is a popular optimization algorithm for parameter estimation in ML models and is particularly effective for large datasets due to its memory efficiency. Unlike first-order optimizers such as Adam or SGD that use only gradient information, L-BFGS incorporates approximate second-order information (the Hessian matrix), resulting in faster convergence and more efficient optimization in GPs.

The GDCPD method addresses two fundamental questions in multivariate CPD: ``Where did the change occur?" and ``Which variables caused the change?". By focusing on the sum of derivatives, GDCPD locates the exact time of change. Furthermore, integrating GPs with ARD, GDCPD identifies which dimensions contributed to these shifts. This makes GDCPD a powerful tool for tackling the complexities of multivariate CPD. Through this comprehensive approach, we improve our ability to diagnose and predict system behavior and enhance our understanding of the underlying mechanisms that drive these changes. With the learned GP model, we can now implement it online to monitor real-time processes.


\section{An Early Warning System}
\label{EW}
This section describes how we use the detected changes and the quantification metrics to develop an early warning system for predicting system failures. These quantification metrics are given in two settings: the offline setting, where the data are given all at once, and the online setting, where the data come sequentially.

\subsection{Mahalanobis Distance and Weighted Mahalanobis Distance}

Upon detecting a process change point, the next step is to quantify the size of this change. We use the WMD to measure the deviation between the pre-change and post-change states of the system, which provide a scalar value that reflects the size of the change.

The MD is a multivariate metric similar to the Euclidean distance but with two significant advantages. First, it is invariant to scale, ensuring that the metric does not change with a rescaling of data. Second, it accounts for the correlations between variables, thereby refines the distance measurement by adjusting for covariances between variables. Suppose we have two independent sample sets, $X^1=\left\{x_1^1, \ldots, x_{d_1}^1\right\}$, and $X^2=\left\{x_1^2, \ldots, x_{d_2}^2\right\}$, with $x_1^1, \ldots, x_{d_1}^1\sim p$, and $x_1^2, \ldots, x_{d_2}^2\sim q$, $p,q \in \Delta_d$ representing distributions before and after a change-point within $d$-dimensional space, respectively. Let $\omega,\zeta \in \Delta_d$ be the empirical distributions of the observations in these two sets, the MD between the two distributions can be expressed as
\begin{equation}
    \text{MD}= \left[(\omega-\zeta)^\top \Sigma^{-1}(\omega-\zeta)\right]^{1/2},
\end{equation}
where $\Sigma^{-1}$ denotes the inverse covariance matrix of $d$ dimensions.

A limitation of MD comes from the fact that this distance is only driven by the variables' covariance structure, not by their practical significance. To address this limitation, WMD fine-tunes MD by incorporating user-defined weights, thereby aligning the distance calculation more closely with the significance of changes across dimensions. In this context, the weights $W$ are determined by the relative importance of derivative changes among variables. Thus, WMD is computed as
\begin{equation}
    \text{WMD}= \left[(\omega-\zeta)^\top W^\top \Sigma^{-1}W(\omega-\zeta)\right]^{1/2},
\end{equation}
which improves the MD by making it sensitive to the importance of changes in these variables.

\subsection{Online Monitoring and Real-Time Alerting}

Leveraging the change-point identification and quantification, we develop an online multivariate process monitoring system. This system uses a fixed sliding window approach over the WMD to continuously evaluate the system's health in real time, enabling the early detection of ill conditions that may lead to system failure.

Unlike the fixed observation duration in offline settings, online observations are received sequentially. The core of our online detection strategy employs a sliding window to dynamically analyze segments of data for changes. For each new observation, the system updates the current window of data, maintaining a fixed window size, that ensures only the most recent observations are considered for analysis. This window slides forward with each new data point, which allows for continuous and real-time evaluation of the system's health status. Given a sequence $\{x_t, t=1,2,\ldots\}$, at each time point $t$, we search over all possible change points ($k<t$) within the fixed window $A$. Consequently, a series of $\{\text{WMD}_\text{W}(k, A)\}_{A\leq k\leq t-A}$ is defined as
\begin{equation}
\label{eq:11}
\begin{aligned}
    \text{WMD}_\text{W}(k,A) = &\big[ \left( \omega(k,A) - \zeta(k-A,A) \right)^\top W^\top \Sigma^{-1} \\
    & \times W\left( \omega(k,A) - \zeta(k-A,A) \right)\big]^{1/2},
\end{aligned}
\end{equation}
where $\omega(k,A)$ and $\zeta(k-A,A)$ represent the empirical distributions of $X$ calculated within a sliding window $[k-A,k+A]$.

The procedure for online CPD defines a stopping time 
\begin{equation}
\mathcal{T}:=\inf \{t:\max_{0\leq k\leq t} \text{WMD}_\text{W}(k,A) \geq b\},
\end{equation}
where $b$ is a pre-specified threshold. As we search over all possible change-points with $k<t$, an alarm is issued if the maximum metric exceeds the threshold (see Algorithm \ref{alg:onlineWMD}).

\begin{algorithm}[t]
\caption{Online WMD}
\label{alg:onlineWMD}
\renewcommand{\algorithmicrequire}{\textbf{Input:}}
\renewcommand{\algorithmicensure}{\textbf{Output:}}

\begin{algorithmic}[1]
    \REQUIRE      \hspace*{\algorithmicindent}{}\\
    Important features: $\tilde d$\\
     GP derivatives $\nabla|\boldsymbol{\mu}_{i,\tilde d}|$\\ 
            Window size: $A$\\
    \ENSURE Online $\text{WMD}_\text{W}$
    \FOR {each one-hour period}
        \FOR {$k$ in range$(0,\text{period}-A+1)$}
            \STATE Compute $\omega = \hat{\mu}(k,k+A)$,\\
            $\zeta = \hat{\mu}(k-A-1,k-1)$
            \STATE Compute $W = \nabla|\boldsymbol{\mu}_{i,\tilde d}|/\sum \nabla|\boldsymbol{\mu}_{i,\tilde d}|$ 
            \STATE Calculate $\text{WMD}_\text{W}$ based on Eq.\eqref{eq:11}
        \ENDFOR
    \ENDFOR
    \RETURN $\text{WMD}_\text{W}$
\end{algorithmic}
\end{algorithm}

The window size must be large enough to capture a meaningful amount of data for analysis but small enough to ensure timely detection. The threshold for triggering an alarm is set as the mean value of calculated WMDs in the offline setting based on historical data.

\subsection{Predicting Remaining Useful Life with LSTM}

\begin{figure}[t!]
\centerline{\includegraphics[width=0.95\textwidth]{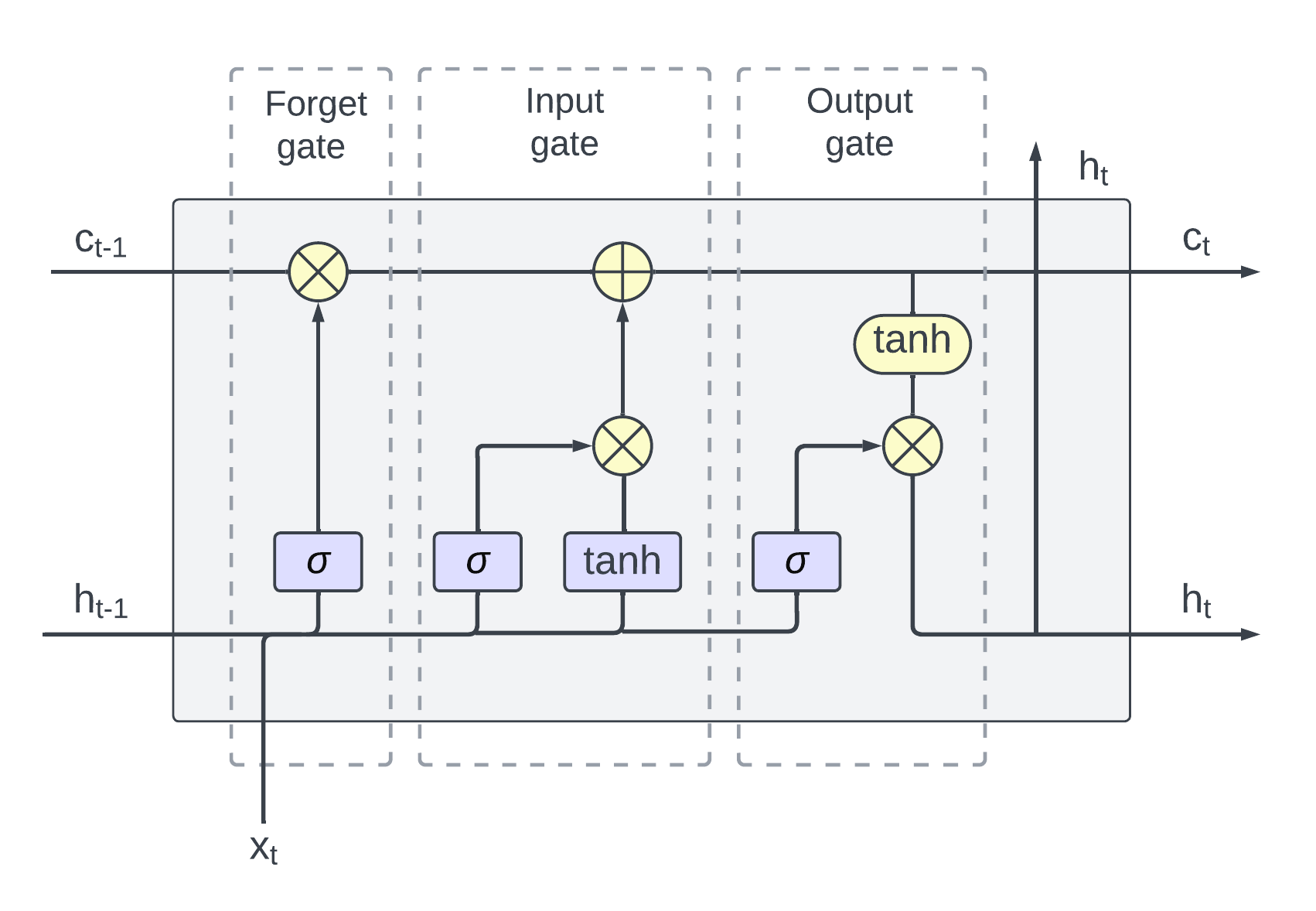}}
\caption{LSTM Architecture.}
\label{fig:lstm}
\end{figure}

LSTM network is a specialized form of Recurrent Neural Networks (RNNs) \cite{yu2019review}. It stands out in analyzing sequential data due to its exceptional ability to capture long-term dependencies. Traditional RNNs often face issues with vanishing and exploding gradients in long sequences.  LSTMs solve this problem with their input, forget, and output gates that can control the information flow. These mechanisms help LSTMs retain relevant historical data and discard irrelevant information, making them particularly suited to predicting the RUL of systems. The architecture of LSTM is shown in Figure~\ref{fig:lstm}. The mathematical expressions can be written as follows,
\begin{equation}
\begin{aligned}
f_t &= \sigma(W_{fh}h_{t-1} + W_{fx}x_t + b_f), \\
i_t &= \sigma(W_{ih}h_{t-1} + W_{ix}x_t + b_i), \\
\tilde{c}_t &= \tanh(W_{ch}h_{t-1} + W_{cx}x_t + b_c), \\
c_t &= f_t \otimes c_{t-1} + i_t \otimes \tilde{c}_t, \\
o_t &= \sigma(W_{oh}h_{t-1} + W_{ox}x_t + b_o), \\
h_t &= o_t \otimes \tanh(c_t),
\end{aligned}
\end{equation}
where $c_t$ denotes the memory state at a given time $t$, and $\tilde{c}_t$ represents the state's updated value. $f_t$, $i_t$, and $o_t$ correspond to the forget gate, input gate, and output gate, respectively. The parameters $W_f$, $W_i$, $W_o$, and $W_c$, along with the $b_f$, $b_i$, $b_o$, and $b_c$ are the weight and biases that need to be learned during training. $\sigma$ is the activation function of gates. The operator ‘$\otimes$’ denotes the pointwise multiplication of two vectors. Note that $x_t$ is the LSTM input at any given time $t$, whereas $h_t$ represents the output from the LSTM cell. When updating the cell state, the forget gate discards information that is irrelevant in the cell state. The input gate decides what useful information can be stored in the cell state. The candidate values $\tilde{c}_t$ are generated by passing $h{t-1}$ and $x_t$ through a tanh function, which produces values in the range $[-1, 1]$. The current state $c_t$ is then updated by combining the previous state and the new candidate values. Finally, the output gate $o_t$ determines what information from the cell state will be output, which produces the hidden state $h_t$ for the next time step.

Our LSTM module of the early warning system utilizes the detected process changes and historical performance data to guide predictive maintenance. To avoid over-fitting and enhance accuracy, we split the entire dataset into training, validation, and testing subsets, and train our LSTM model with the features from GDCPD and WMD after removing the observations during the unstable period immediately following the last system breakdown. Then, we use this model to predict the RUL on the testing dataset after an early warning alarm is issued.


\section{Experimental Study}
\label{experiment}

\subsection{Dataset}

The dataset\footnote{The data is available at \cite{ranjan2018dataset}.}  is provided by a pulp-and-paper mill and is aimed at addressing a critical challenge in paper manufacturing: predicting and preventing paper breaks during the continuous rolling process \cite{ranjan2018dataset}. These breaks not only halt production but also lead to significant financial losses due to the downtime required for problem resolution and process restart. This dataset captures data from sensors placed across the paper machine, which measure both raw materials (e.g., amount of pulp fiber, chemicals, etc.) and process variables (e.g., blade type, couch vacuum, rotor speed, etc.). Specific sensor descriptions are omitted to maintain data anonymity. The data consists of 18,398 time stamps with a two-minute interval between consecutive time stamps and includes a binary response variable indicating whether a break occurred. A total of 124 paper breakages have been identified.

Before inputting the data into the analysis framework, we perform preprocessing steps to ensure data quality and consistency. We standardize all sensor measurements to ensure uniform scale and reliability across variables, and missing data points are handled using linear interpolation to fill gaps. We focus on the data within the critical one-hour window preceding an early alarm to provide actionable early warnings. Data collected immediately after a system breakdown is excluded from consideration because the system is unstable during the restart phase, which could skew the analysis and lead to unreliable predictions. The dataset is divided into training, validation, and test subsets, allocated 60\%, 20\%, and 20\% of the original data, respectively.


\subsection{Results and Discussion}

\begin{figure}[t!]
\centerline{\includegraphics[width=0.75\textwidth]{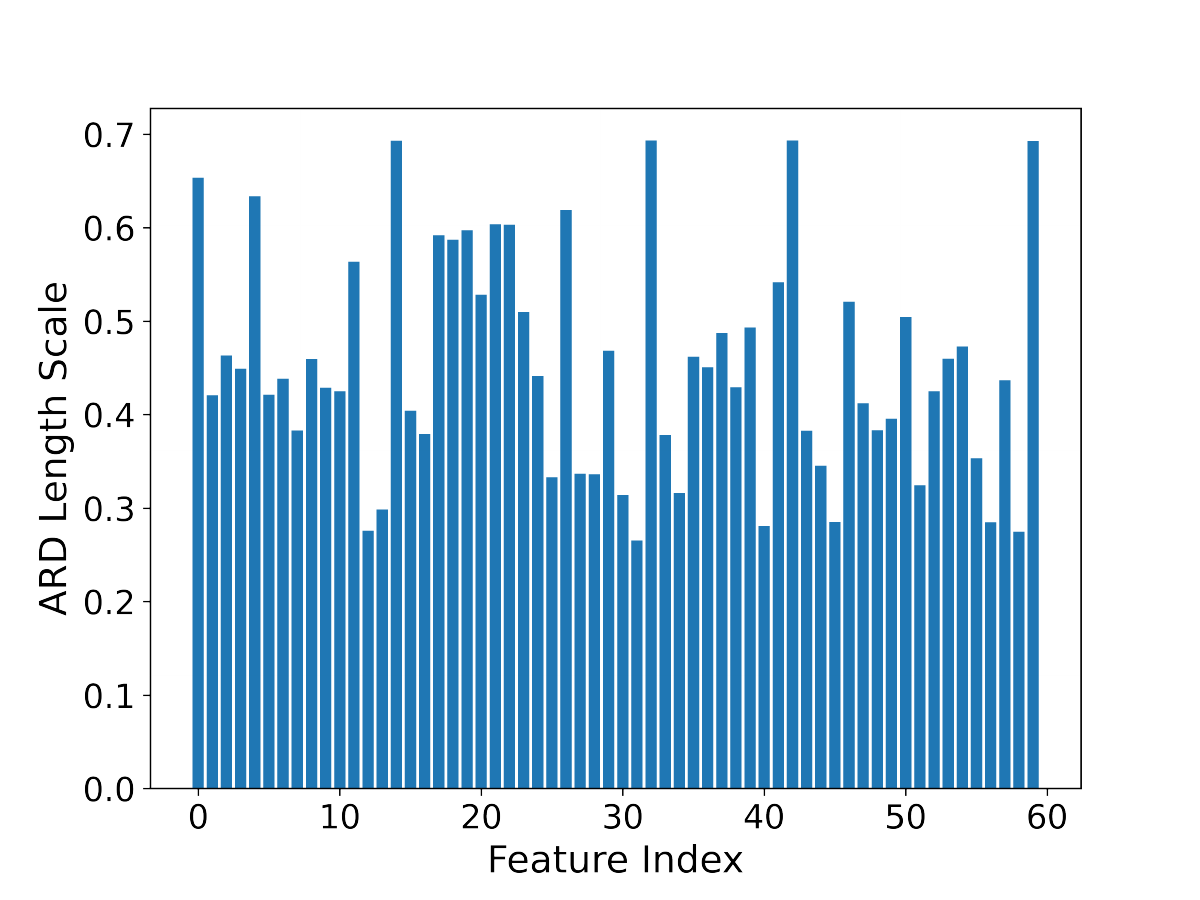}}
\caption{ARD Length-scales of Features.}
\label{fig:ard_lengthscales}
\end{figure}

\begin{figure}[h!]
\centerline{\includegraphics[width=1.2\textwidth]{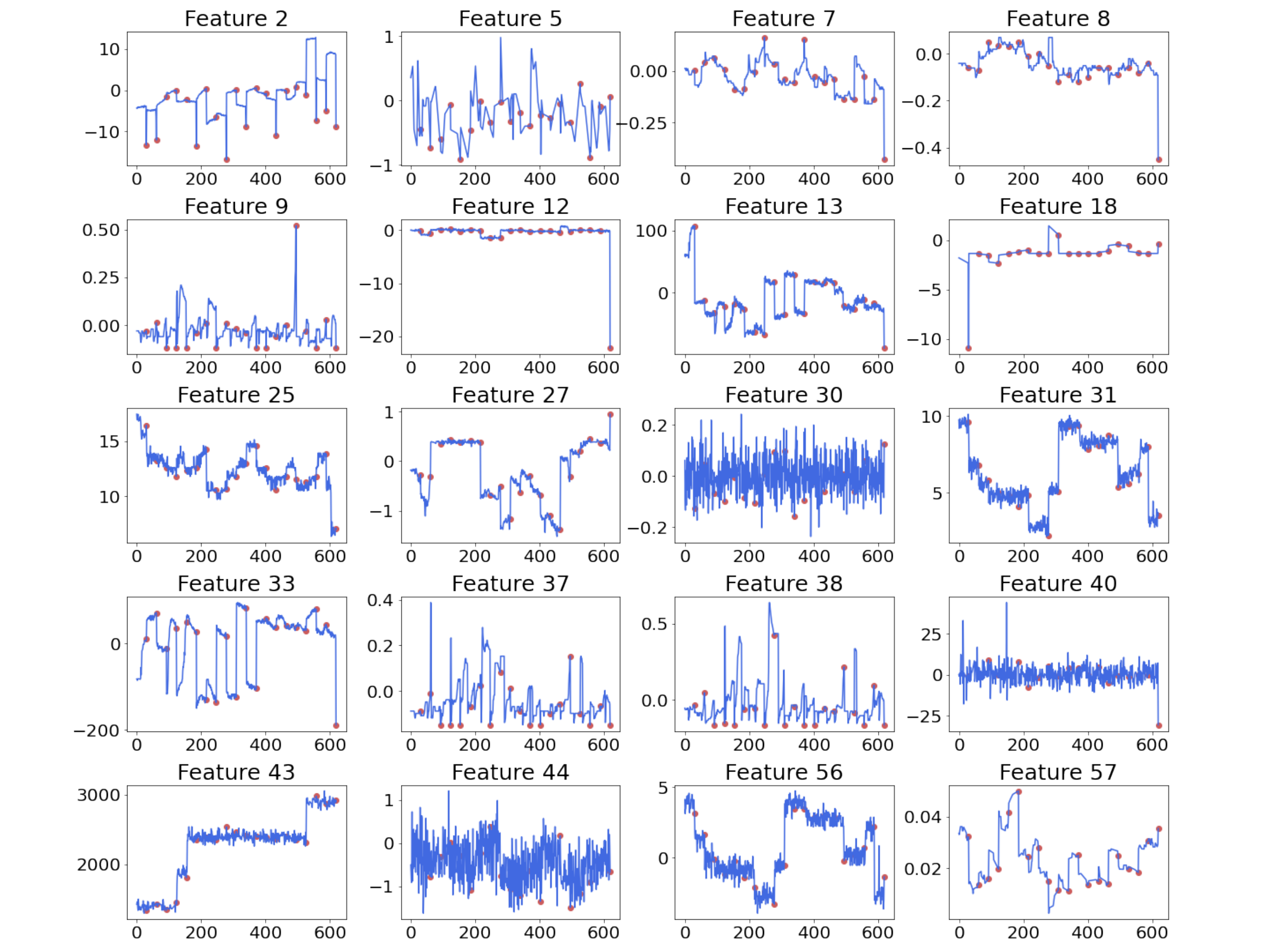}}
\caption{
Selected Features with Small Length-Scale. These selected features were identified by the ARD mechanism of multi-input GP classification model. The red dots indicate the times of system breakdowns, while the feature values over a one-hour period preceding system breakdown are plotted by blue line segments. Notably, certain features show abrupt changes or significant fluctuations prior to breakdowns, which suggests a potential predictive relationship between feature changes and system failures.}
\label{fig:ard_ls_0.45}
\end{figure}

\begin{figure}[h!]
\centerline{\includegraphics[width=1.0\textwidth]{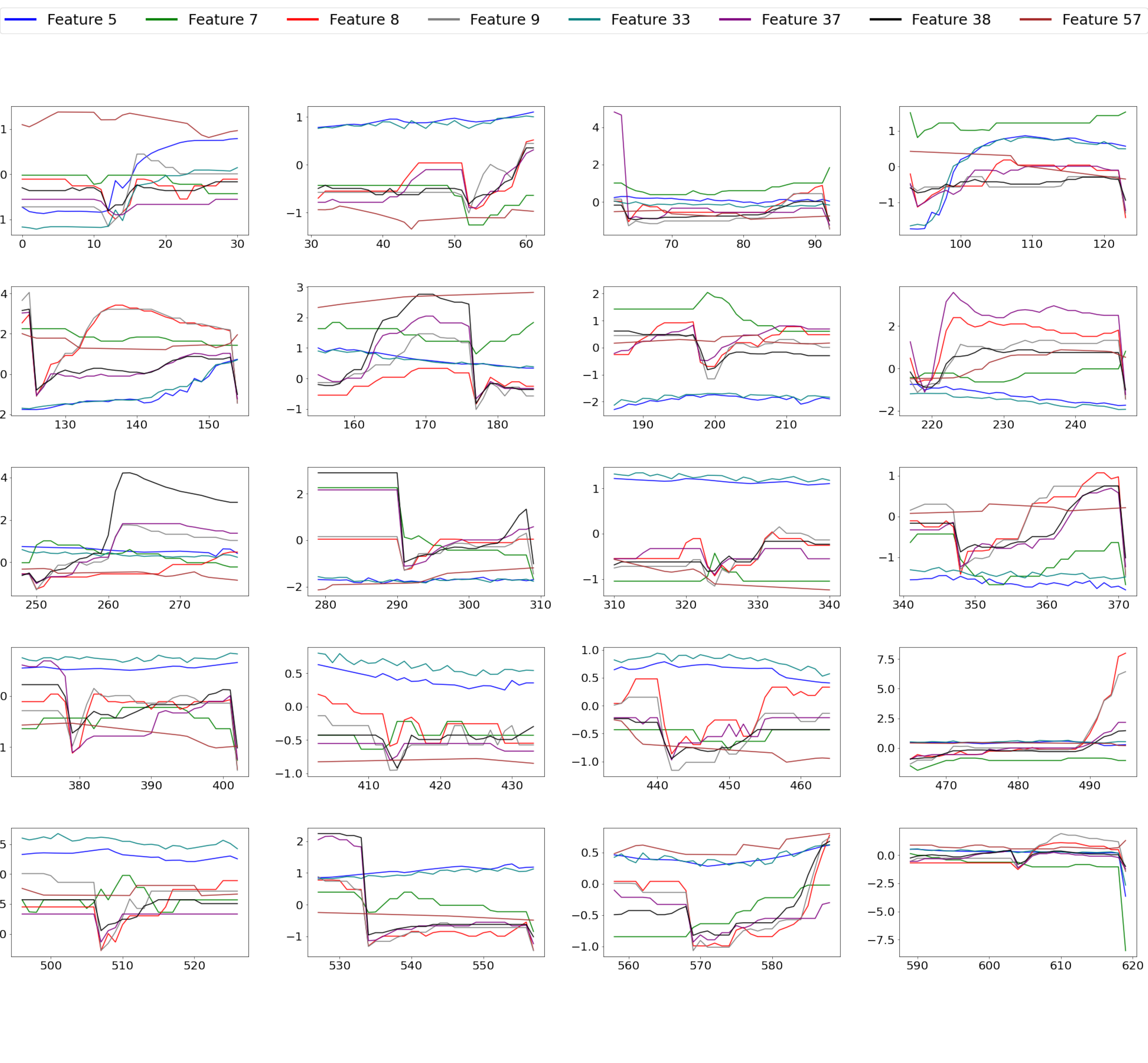}}
\vspace{-0.3in}
\caption{Hourly Pre-Shutdown Feature Behavior Analysis. This figure presents the behavior of key features within the one-hour window before manufacturing shutdown events. These features were identified through the multivariate GP model with an ARD length-scale below 0.45, which indicates high relevance. Each subplot corresponds to one of the first 20 shutdown events in the dataset, with data points plotted at 2-minute intervals. The consistent patterns and abrupt changes observed across multiple shutdowns indicates the potential of these features as early indicators for system failures.
}
\label{fig:features}
\end{figure}

GDCPD is employed in this experiment and the corresponding ARD length-scales are illustrated in Figure~\ref{fig:ard_lengthscales}. We set 0.45 as the threshold for identifying the most relevant dimensions. This threshold indicates that features with length-scales below this value are considered highly relevant since a smaller length-scale value means an increasing model sensitivity to the variation in that specific feature. But note that this sensitivity does not necessarily imply abrupt changes in the feature values themselves.  Instead, the ARD length-scale reflects the degree to which fluctuations in a feature correspond to the changes in the response variable. And that provides the importance of each feature to system behavior.

\begin{figure}[t!]
\centerline{\includegraphics[width=1.2\textwidth]{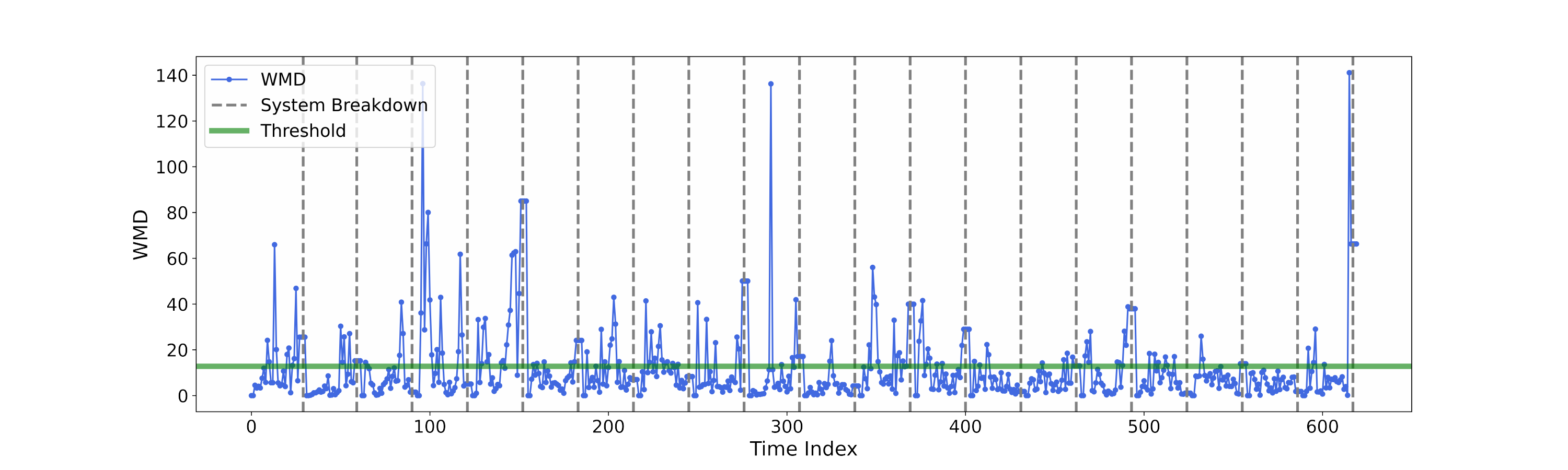}}
\caption{The WMD Values for the Online Monitoring of the First 20 System Breakdowns. The grey dashed lines represent one hour before each system shutdown. The detection threshold is set at 12.8045.}
\label{fig:WMD_online}
\end{figure}

When analyzing system failures, we observed that abrupt changes in certain feature values often coincide with system breakdowns. Figure~\ref{fig:ard_ls_0.45} illustrates this phenomenon, some features exhibit significant fluctuations or sudden shifts prior to a failure event. In some cases, system failures cause sudden changes in feature values rather than the feature changes leading to failures. However, in other instances, significant deviations in key feature values can indeed result in an immediate system shutdown. These interactions between feature behaviors and system failures make it challenging to establish a clear cause-and-effect relationship. To address this issue and identify features that can provide early warning signs, we utilize the chronological sequence of events; i.e., we will only consider the features whose changes occur well before the early warning within the one-hour time window.

By examining the data, we can identify the features that exhibit abrupt shifts within the critical one-hour window before a warning alarm, as illustrated in Figure~\ref{fig:features}. We applied the GDCPD method to analyze the trends and behaviors of these features, which led us to select eight features that consistently demonstrated significant mean shifts prior to system failures, indicating their potential as early warning indicators. By monitoring these features, it may be possible to provide early warnings and take preventive actions to avoid breakdowns. To quantify the changes, we calculated the WMD for each one-hour data segment. At this offline step, we assume that there is only one change point within each time window. The average WMD value across the training set was calculated to be 12.8045, which we adopted as the threshold for triggering an early warning.

\begin{figure}[t!]
\centerline{\includegraphics[width=0.69\textwidth]{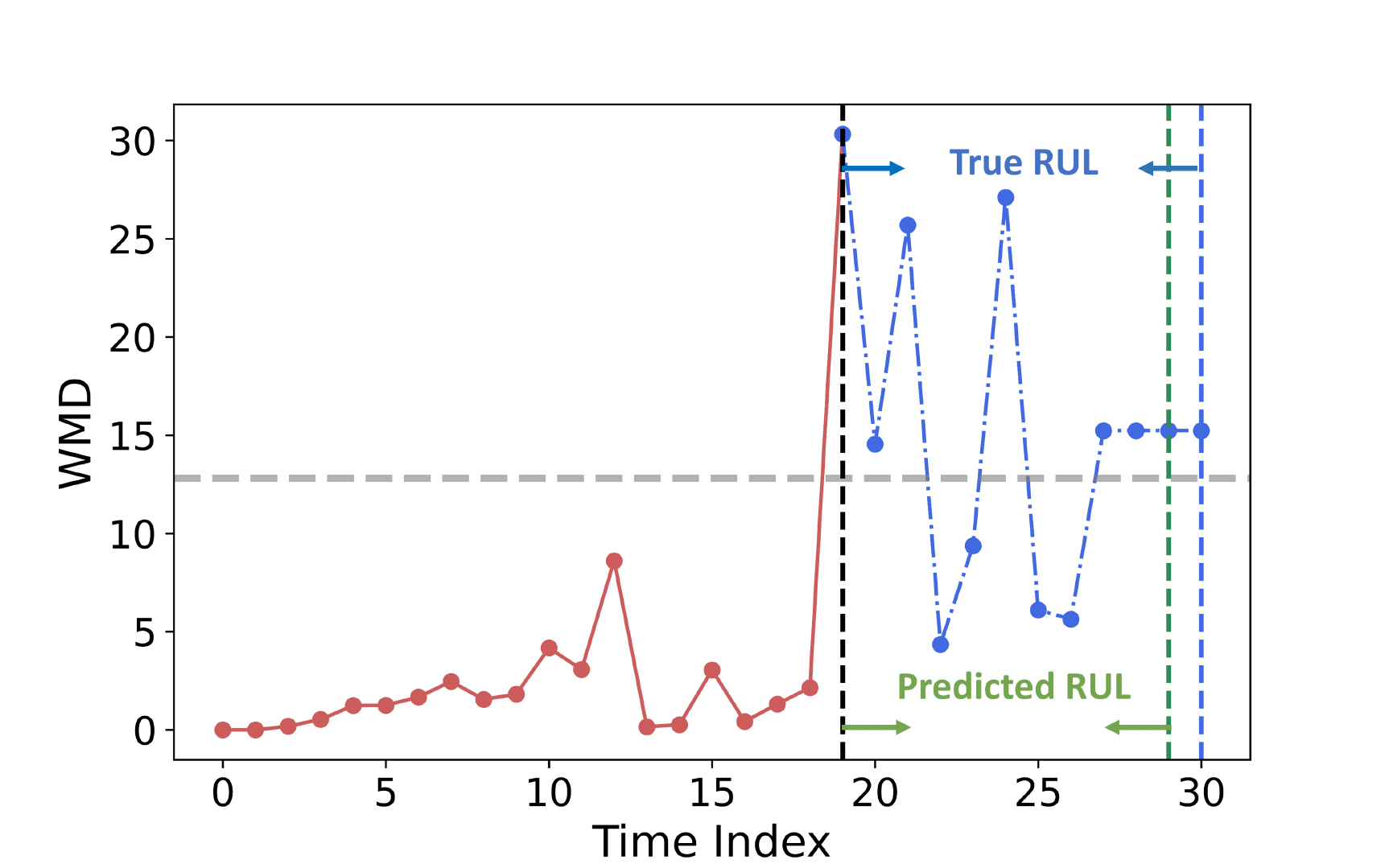}}
\caption{The RUL Prediction. The WMD data starting one hour before early warning are used to train LSTM models. The vertical black dashed line indicates the time when an early warning is issued.}
\label{fig:RUL}
\end{figure}

In our online system monitoring, a sliding window technique was employed to track the WMD with each window spanning a six-minute interval (illustrated in Figure~\ref{fig:WMD_online}). This strategy ensures early warnings are triggered at least six minutes before a potential system failure. High WMD readings that trigger alerts typically occur when the sliding window passes over an emerging change multiple times. When a system breakdown did happen, immediately after repair, the system could be unstable, which would result in elevated WMD values at the beginning; therefore, these immediate-after-setup values should be ignored. We begin monitoring WMDs after a certain period following repairs to ensure the system has stabilized and to avoid false alarms from initial fluctuations. Similarly, high WMD values at the end of the monitoring period often signal the onset of a breakdown, capturing the system's final moments before failure. 

We used the eight critical features, along with the corresponding time stamps and real-time WMD measurements, as predictors. The LSTM model was trained on the one-hour window data sequence leading up to the first exceed of the WMD threshold, as depicted in Figure~\ref{fig:RUL}. This ensures that the model learns from the most relevant and immediate precursors to system failures. Once the model is adapted to these initial data from the breakdown cycle, it begins making real-time RUL predictions for the subsequent data points, and the model is continuously updated with new data coming in. For testing, before making predictions, we employed the one-hour data segment from each testing breakdown cycle to calibrate the model. This calibration is not for learning from the test data but to adjust the model to the specific conditions of each test cycle. After adaptation, the model predicted the remaining of the cycle until the RUL reached 10 minutes. For comparison experiments, the model used data from the beginning of one hour of system operation to perform a one-time prediction.

Our model utilizes a three-layer LSTM architecture, each layer equipped with 100 hidden neurons, designed to estimate RUL effectively. To prevent overfitting, we have incorporated dropout layers between each LSTM layer. The dropout rate is set at 0.2, meaning 20\% of the neurons are randomly dropped out during training, which helps ensure that the model does not rely too heavily on randomly dropped neurons and thereby improves generalizability. Following the LSTM layers, the model combined a fully connected output layer that translates the learned features into a predicted RUL value. We trained the model over 30 epochs using the Adam optimizer.

\begin{table}
    \centering
    \begin{tabular}{|c|c|c|c}
    \hline
        Methods & RMSE & SF \\
        \hline
        LSTM \cite{ zheng2017long} &0.3634  & 0.4123 \\
        Bi-LSTM \cite{huang2019bidirectional} & 0.4327 & 0.3256  \\
        Proposed method (CPD+LSTM) &  \textbf{0.1560 }&  \textbf{0.1206}\\
    \hline
    \end{tabular}
    \caption{Performance Comparison between the Proposed Method and Existing Algorithms (best in bold).}
    \label{tab:comparison}
\end{table}

To evaluate the performance of RUL estimation models, we employed two benchmark metrics: Root Mean Square Error (RMSE) and the Score Function (SF). The SF \cite{saxena2008damage} is asymmetric around the true time of failures and has heavier penalties for overestimating the RUL because it could lead to unnecessary delays in maintenance and potential system failures. Our approach outperforms other LSTM methods, as shown in Table \ref{tab:comparison}, demonstrating its superior accuracy and robustness in RUL prediction. This confirms the effectiveness of our model in practical predictive maintenance applications. 


\section{Conclusion}
\label{Conclusion}

In summary, we develop an innovative approach to system health monitoring and early warning of failure in this paper. GDCPD enhances the ability to detect abrupt changes in multivariate processes. The offline WMD analysis establishes a reliable threshold for detecting system changes and introduces a metric for quantifying changes in online settings. Testing our methodology on a real-world system has demonstrated its effectiveness in accurately predicting system failures and enabling proactive maintenance strategies. By integrating GDCPD for detecting changes, WMD for threshold setting, and LSTM for RUL prediction, our approach has significantly advanced the system health monitoring and predictive maintenance knowledge base. It enhances the reliability and efficiency of manufacturing processes.

\section{Acknowledgements}
This work was supported by the National Science Foundation [award number 2134409].


\appendix
\section{Posterior Distribution of GP Derivatives}
\label{A}
Suppose the posterior distribution at a particular test point $\boldsymbol{x}^{\prime}$, with training inputs $\boldsymbol{x}$, training targets $\boldsymbol{y}$, and hyperparameters $\boldsymbol{\theta}$ be,
\begin{equation}
p(f|\boldsymbol{y}, \boldsymbol{x}, \boldsymbol{x}^{\prime},\boldsymbol{\theta}) \sim \mathcal{N}(\boldsymbol{\mu}, \boldsymbol{k}).
\end{equation}

We represent the unknown GP function value $f$ as a combination of its mean and stochastic deviation term $\epsilon$, expressed as follows:
\begin{equation}
\begin{aligned}
f(\boldsymbol{x}^{\prime}) &= \bar{f}(\boldsymbol{x}^\prime) + \epsilon^{\prime},\\
f(\boldsymbol{x}^{\prime}+ \boldsymbol{\delta}_i) &= \bar{f}(\boldsymbol{x}^\prime + \boldsymbol{\delta}_i) + \epsilon_{\delta_i}.
\end{aligned}
\end{equation}

For convenience,  we use the shorthand that,
\begin{equation}
k_{\epsilon^{\prime}\epsilon^{\prime}} \triangleq k(\epsilon^{\prime}, \epsilon^{\prime}), \quad
\boldsymbol{k}_{\epsilon^{\prime}} \triangleq k(\boldsymbol{x}, \epsilon^{\prime}), \quad
K \triangleq K(\boldsymbol{x}, \boldsymbol{x}).
\end{equation}

Here, the joint distribution of $\epsilon^{\prime}$ and  $\epsilon_{\delta_i}$ is modeled as a zero-mean Gaussian:
\begin{equation}
p(\epsilon^{\prime}, \epsilon_{\delta_i}) = \mathcal{N} \left( \boldsymbol{0}, 
\begin{bmatrix}
k_{\epsilon^{\prime}\epsilon^{\prime}} - \boldsymbol{k}_{\epsilon^{\prime}}^\top K^{-1} \boldsymbol{k}_{\epsilon^{\prime}} 
& k_{\epsilon^{\prime}\epsilon_{\delta_i}} - \boldsymbol{k}_{\epsilon^{\prime}}^\top K^{-1} \boldsymbol{k}_{\epsilon_{\delta_i}} \\
k_{\epsilon_{\delta_i}\epsilon^{\prime}} - \boldsymbol{k}_{\epsilon_{\delta_i}}^\top K^{-1} \boldsymbol{k}_{\epsilon^{\prime}} 
& k_{\epsilon_{\delta_i}\epsilon_{\delta_i}} - \boldsymbol{k}_{\epsilon_{\delta_i}}^\top K^{-1} \boldsymbol{k}_{\epsilon_{\delta_i}}
\end{bmatrix} \right),
\end{equation}
where $\boldsymbol{\delta}_i$ is a perturbation vector on $\boldsymbol{x}^{\prime}$, defined as:
\begin{equation}
    \boldsymbol{\delta}_i = [0_1, \ldots, 0_{i-1}, \delta, 0_{i+1}, \ldots, 0_d]^\top.
\end{equation}

Given $f(\boldsymbol{x}^{\prime})$ as a scalar, its derivatives $ \frac{\partial f(\boldsymbol{x}^{\prime})}{\partial \boldsymbol{x}^{\prime}} $ is a vector in d-dimension space with its $i$-th component being the partial derivative with respect to $\boldsymbol{x}^{\prime}_{(i)}$:
\begin{equation}
\label{eq:A5}
\begin{aligned}
\frac{\partial f}{\partial x^{\prime}_{(i)}} &= \lim_{\delta \to 0} \frac{f(\boldsymbol{x}^{\prime}+ \boldsymbol{\delta}_i) - f(\boldsymbol{x}^{\prime})}{x^\prime_{(i)}+\delta-x^\prime_{(i)}} \\
&= \lim_{\delta \to 0} \frac{\bar{f}(\boldsymbol{x}^\prime + \boldsymbol{\delta}_i) + \epsilon_{\delta_i} - \bar{f}(\boldsymbol{x}^\prime) - \epsilon^{\prime}}{\delta} \\
&= \lim_{\delta \to 0} \frac{\bar{f}(\boldsymbol{x}^\prime + \boldsymbol{\delta}_i) - \bar{f}(\boldsymbol{x}^\prime)}{\delta} + 
\lim_{\delta \to 0} \frac{\epsilon_{\delta_i} - \epsilon^{\prime}}{\delta} \\
&= \frac{\partial \bar{f}}{\partial x^{\prime}_{(i)}} + \lim_{\delta \to 0} \frac{\epsilon_{\delta_i} - \epsilon^{\prime}}{\delta}.
\end{aligned}
\end{equation}

The first term in eq.~\ref{eq:A5} is the derivative of the mean function, and the second term is the Gaussian term. The expectation of $\frac{\partial f}{\partial x^{\prime}_{(i)}}$ aligns with the derivative of the mean function since differentiation and expectation are both linear operators. Then the posterior mean function for a GP is:
\begin{equation}
\begin{aligned}
   \bar{f} &= \boldsymbol{k}(\boldsymbol{x}, \boldsymbol{x}^{\prime}) K^{-1} \boldsymbol{y},\\
\mathbb{E}_f \left[ \frac{\partial f}{\partial x^{\prime}_{(i)}}\right]^\top &= \frac{\partial \boldsymbol{k}(\boldsymbol{x}, \boldsymbol{x}^{\prime}) }{\partial \boldsymbol{x}^{\prime}_{(i)}}K^{-1} \boldsymbol{y}.
\end{aligned}\end{equation}

To evaluate the variance of the derivative, we consider the second term in eq.~\ref{eq:A5}:

\begin{equation}
\begin{aligned}
\mathbb{V}_{f} \left[ \frac{\partial f}{\partial x^{\prime}_{(i)}} \right] =& \mathbb{V} \left[ \lim_{\delta \to 0} \frac{\epsilon_{\delta_i} - \epsilon^{\prime}}{\delta} \right]\\
=& \lim_{\delta \to 0} \frac{1}{\delta^2} \left( \mathbb{V}[\epsilon_{\delta_i}] + \mathbb{V}[\epsilon^{\prime}] - \mathbb{C}[\epsilon_{\delta_i}, \epsilon^{\prime}] - \mathbb{C}[\epsilon^{\prime}, \epsilon_{\delta_i}] \right)\\
=& \lim_{\delta \to 0} \frac{1}{\delta^2} \left( k_{\epsilon_{\delta_i}\epsilon_{\delta_i}} - \boldsymbol{k}_{\epsilon_{\delta_i}}^\top K^{-1} \boldsymbol{k}_{\epsilon_{\delta_i}} 
+ k_{\epsilon^{\prime}\epsilon^{\prime}} - \boldsymbol{k}_{\epsilon^{\prime}}^\top K^{-1} \boldsymbol{k}_{\epsilon^{\prime}}   \right.\\
&-\left.(\boldsymbol{k}_{\epsilon_{\delta_i}}-\boldsymbol{k}_{\epsilon^{\prime}})^T K^{-1} (\boldsymbol{k}_{\epsilon_{\delta_i}}-\boldsymbol{k}_{\epsilon^{\prime}})\right)\\
=& \frac{\partial^2 k(\boldsymbol{x}^{\prime}, \boldsymbol{x}^{\prime})}{\partial x^{\prime}_{(i)} \partial x^{\prime}_{(i)}} - \frac{\partial \boldsymbol{k}(\boldsymbol{x}^{\prime}, \boldsymbol{x})}{\partial x^{\prime}_{(i)}} K^{-1} \frac{\partial \boldsymbol{k}(\boldsymbol{x}, \boldsymbol{x}^{\prime})}{\partial x^{\prime}_{(i)}}.
\end{aligned}
\end{equation}

Thus we can find the covariance matrix,
\begin{equation}
\mathbb{C}_{f} \left[ \frac{\partial f}{\partial x^{\prime}_{(i)}}, \frac{\partial f}{\partial x^{\prime}_{(j)}} \right] = \frac{\partial^2 k(\boldsymbol{x}^{\prime},\boldsymbol{x}^{\prime})}{\partial x^{\prime}_{(i)} \partial x^{\prime}_{(j)}} - \frac{\partial \boldsymbol{k}(\boldsymbol{x}^{\prime}, \boldsymbol{x})}{\partial x^{\prime}_{(i)}} K^{-1} \frac{\partial \boldsymbol{k}(\boldsymbol{x}, \boldsymbol{x}^{\prime})}{\partial x^{\prime}_{(j)}}.
\end{equation}

Therefore, the posterior distribution of GP derivatives is:
\begin{equation}
\begin{aligned}
p \bigg( \frac{\partial f}{\partial \boldsymbol{x}^{\prime}} \bigg|  & \boldsymbol{x}^{\prime}, \boldsymbol{x}, \boldsymbol{y}, \boldsymbol{\theta}  \bigg) \\
&= \mathcal{N} \left( \frac{\partial \boldsymbol{k}(\boldsymbol{x}, \boldsymbol{x}^{\prime}) }{\partial \boldsymbol{x}^{\prime}}K^{-1} \boldsymbol{y}, \; \frac{\partial^2 k(\boldsymbol{x}^{\prime},\boldsymbol{x}^{\prime})}{\partial \boldsymbol{x}^{\prime} \partial \boldsymbol{x}^{\prime\top}} - \frac{\partial \boldsymbol{k}(\boldsymbol{x}^{\prime}, \boldsymbol{x})}{\partial \boldsymbol{x}^{\prime}} K^{-1} \frac{\partial \boldsymbol{k}(\boldsymbol{x}, \boldsymbol{x}^{\prime})}{\partial \boldsymbol{x}^{\prime}} \right).
\end{aligned}
\end{equation}

\section{Derivatives of RBF Kernel}
\label{B}
We define the RBF kernel function as:
\begin{equation}
k(\boldsymbol{x},\boldsymbol{x}^{\prime})=
\sigma^2 \exp \left(-\frac{1}{2}\left(\boldsymbol{x}-\boldsymbol{x}^{\prime}\right)^\top 
 \Lambda^{-1}\left(\boldsymbol{x}-\boldsymbol{x}^{\prime}\right)\right).
 \end{equation}
The hyperparameters consist of the signal variance $\sigma^2$ and a length-scale for each dimension, denoted as $\{\ell_i\}_{i=1}^D$. These length-scale squares are the entries for a diagonal matrix $\Lambda$ of dimensions $D \times D$,
\begin{equation}
\Lambda = \begin{bmatrix}
\ell_1^2 & 0 & \cdots & 0 \\
0 & \ell_2^2 & \cdots & 0 \\
\vdots & \vdots & \ddots & \vdots \\
0 & 0 & \cdots & \ell_D^2
\end{bmatrix}.
\end{equation}

The derivative of RBF kernel with respect to $\boldsymbol{x}^{\prime}$ is:
\begin{equation}
\begin{aligned}
\frac{\partial k\left(\boldsymbol{x}, \boldsymbol{x}^{\prime}\right)}{\partial \boldsymbol{x}^{\prime}} & =\frac{\partial}{\partial \boldsymbol{x}^{\prime}}\left\{\sigma^2 \exp \left(-\frac{1}{2}\left(\boldsymbol{x}-\boldsymbol{x}^{\prime}\right)^\top 
 \Lambda^{-1}\left(\boldsymbol{x}-\boldsymbol{x}^{\prime}\right)\right)\right\} \\
& =\frac{\partial}{\partial \boldsymbol{x}^{\prime}}\left\{-\frac{1}{2}\left(\boldsymbol{x}-\boldsymbol{x}^{\prime}\right)^\top \Lambda^{-1}\left(\boldsymbol{x}-\boldsymbol{x}^{\prime}\right)\right\} k\left(\boldsymbol{x}, \boldsymbol{x}^{\prime}\right) \\
& =-\frac{1}{2} \frac{\partial}{\partial \boldsymbol{x}^{\prime}}\left\{-\boldsymbol{x}^{\prime\top} \Lambda^{-1} \boldsymbol{x}-\boldsymbol{x}^\top \Lambda^{-1} \boldsymbol{x}^{\prime}+\boldsymbol{x}^{\prime\top} \Lambda^{-1} \boldsymbol{x}^{\prime}\right\} k\left(\boldsymbol{x}, \boldsymbol{x}^{\prime}\right) \\
& =-\frac{1}{2}\left(-2 \Lambda^{-1} \boldsymbol{x}+2 \Lambda^{-1} \boldsymbol{x}^{\prime}\right) k\left(\boldsymbol{x}, \boldsymbol{x}^{\prime}\right) \\
& =\Lambda^{-1}\left(\boldsymbol{x}-\boldsymbol{x}^{\prime}\right) k\left(\boldsymbol{x}, \boldsymbol{x}^{\prime}\right).
\end{aligned}
\end{equation}
The mixed derivative is:
\begin{equation}
\begin{aligned}
\frac{\partial^2 k\left(\boldsymbol{x}, \boldsymbol{x}^{\prime}\right)}{\partial \boldsymbol{x} \partial \boldsymbol{x}^{\prime}} & =\frac{\partial}{\partial \boldsymbol{x}}\left\{\Lambda^{-1}\left(\boldsymbol{x}-\boldsymbol{x}^{\prime}\right) k\left(\boldsymbol{x}, \boldsymbol{x}^{\prime}\right)\right\} \\
& =\Lambda^{-1}\left(\delta-\left(\boldsymbol{x}-\boldsymbol{x}^{\prime}\right)\left(\boldsymbol{x}-\boldsymbol{x}^{\prime}\right)^T \Lambda^{-1}\right) k\left(\boldsymbol{x}, \boldsymbol{x}^{\prime}\right).
\end{aligned}
\end{equation}

\section{Additional Experiments}
\label{C}
\subsection{Simulation Experiments}

\begin{figure}[h]
  \centering
  \begin{subfigure}[b]{0.24\linewidth}
    \includegraphics[width=\linewidth]{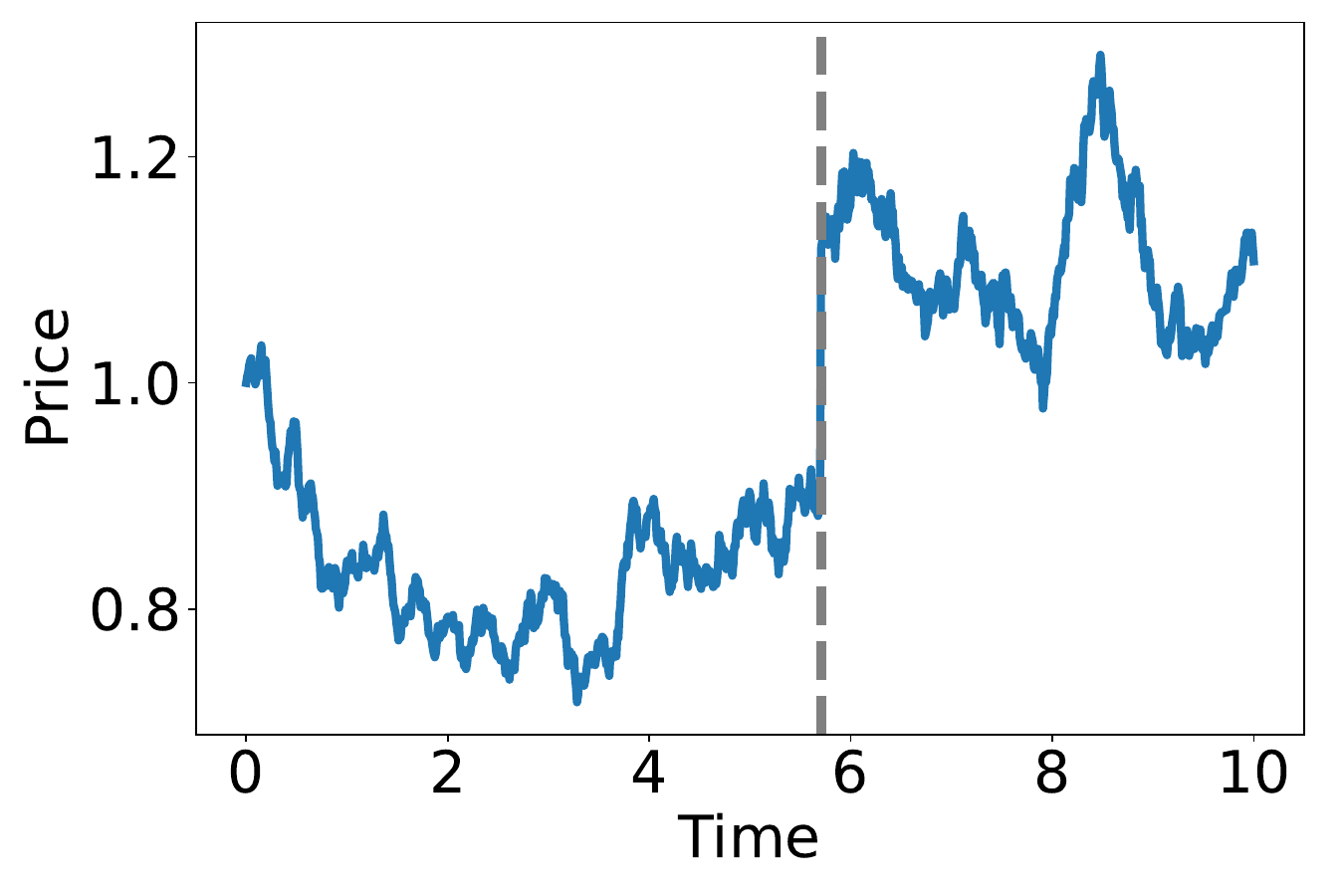}
     \caption{$\text{MJD}_{t_{no}}$}
  \end{subfigure}
  \begin{subfigure}[b]{0.24\linewidth}
    \includegraphics[width=\linewidth]{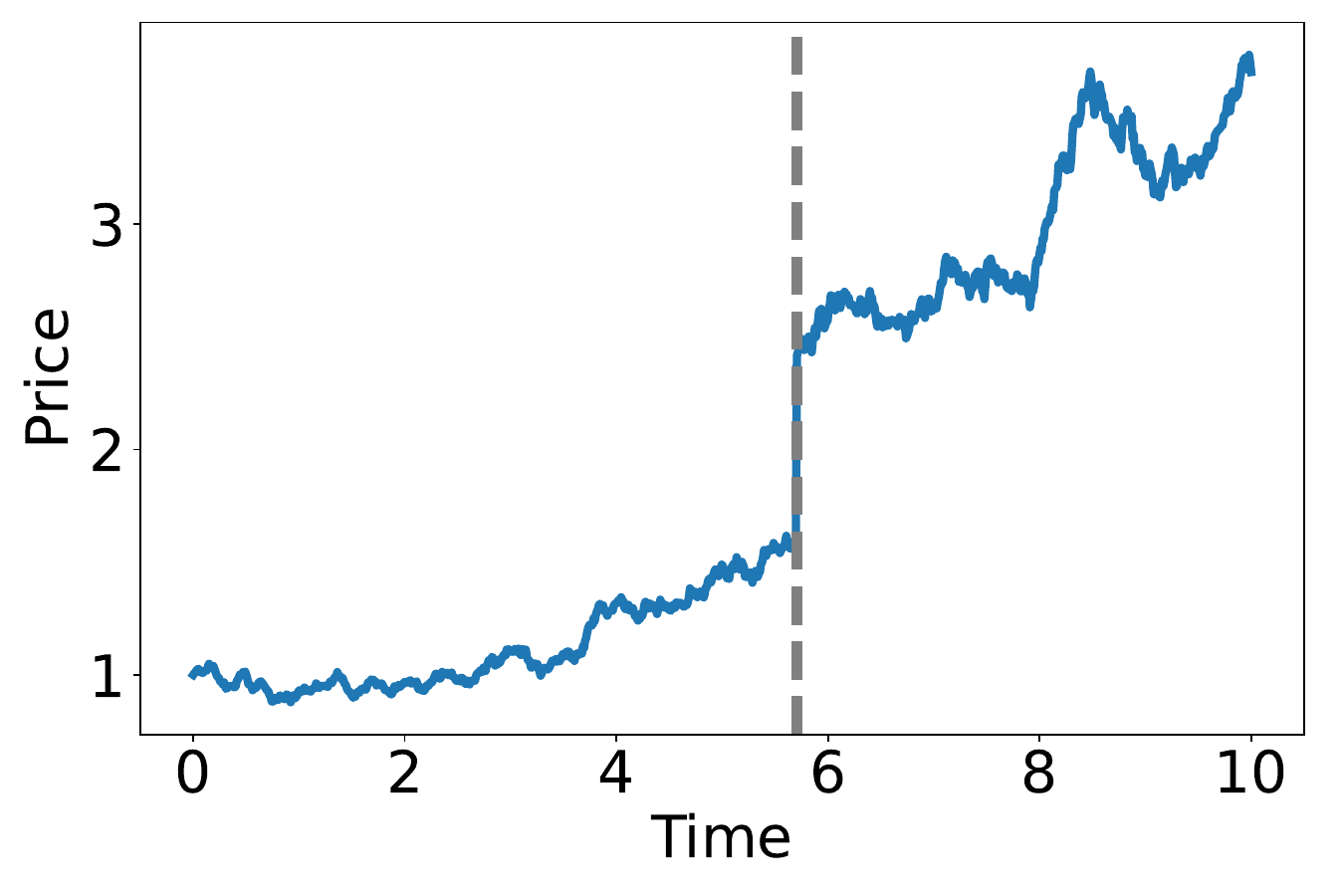}
    \caption{$\text{MJD}_{t_{up}}$}
  \end{subfigure}
  \begin{subfigure}[b]{0.24\linewidth}
    \includegraphics[width=\linewidth]{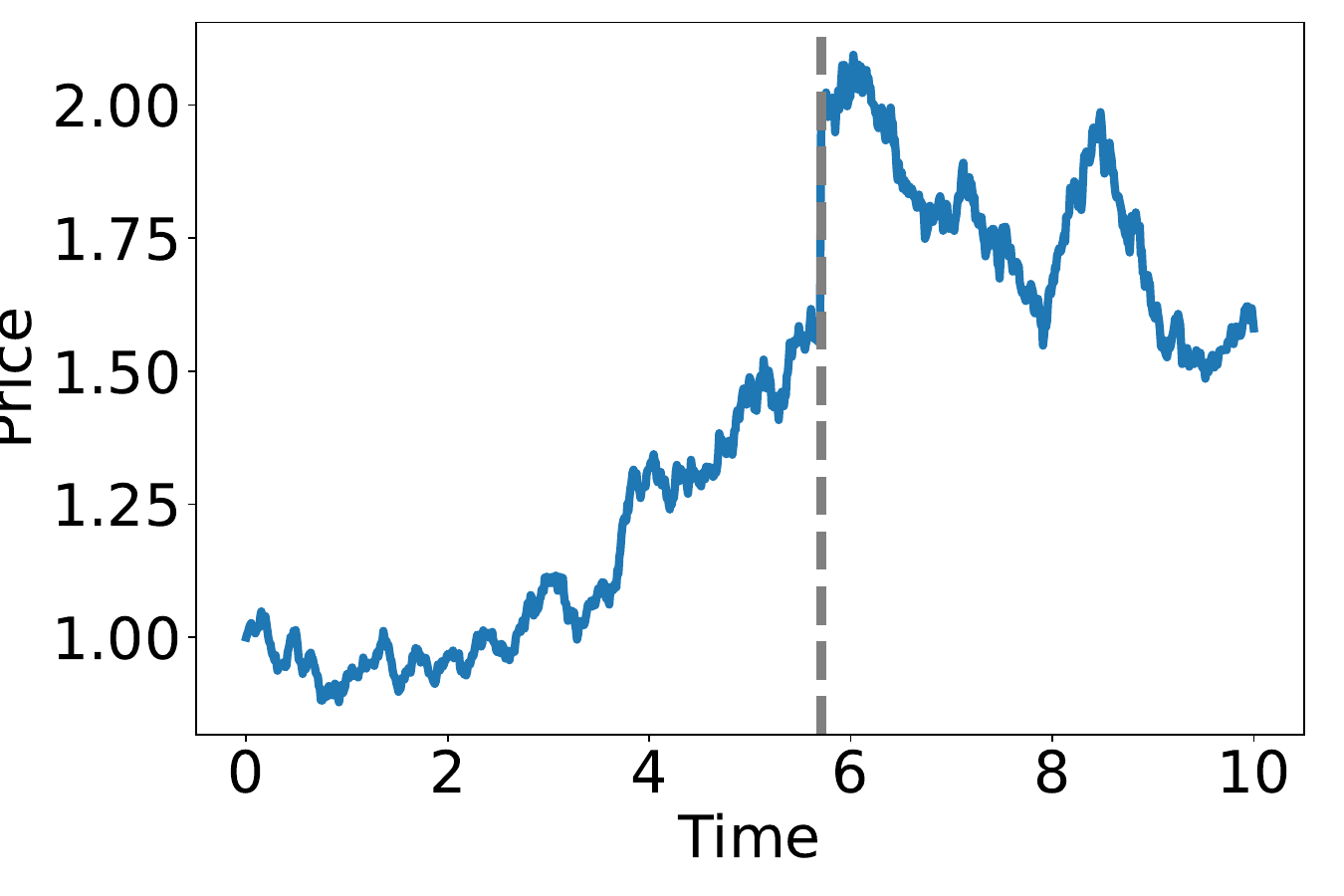}
    \caption{$\text{MJD}_{t_{inv}}$}
  \end{subfigure}
  \begin{subfigure}[b]{0.24\linewidth}
    \includegraphics[width=\linewidth]{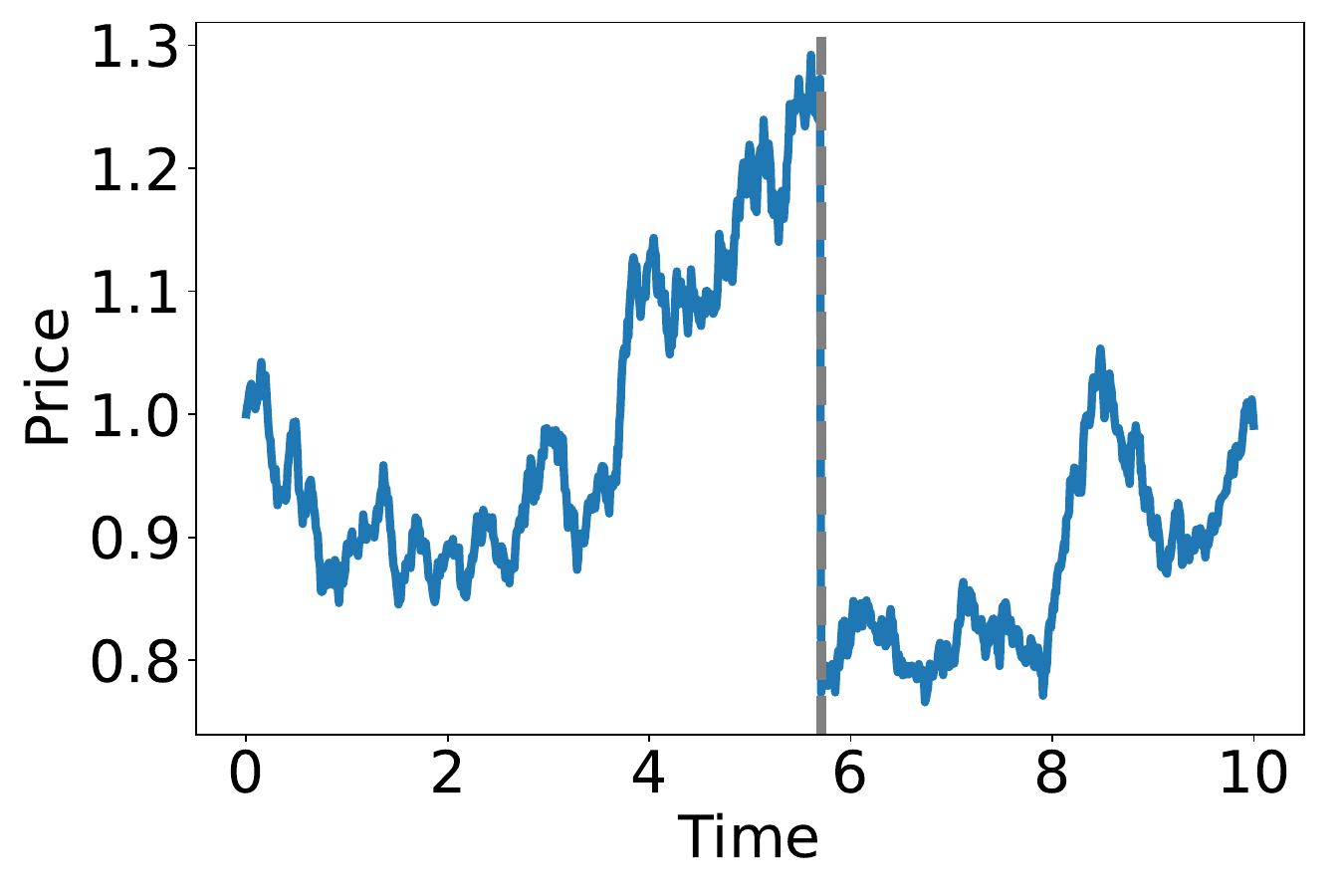}
    \caption{$\text{MJD}_{t_{down}}$}
  \end{subfigure}
    \begin{subfigure}[b]{0.24\linewidth}
    \includegraphics[width=\linewidth]{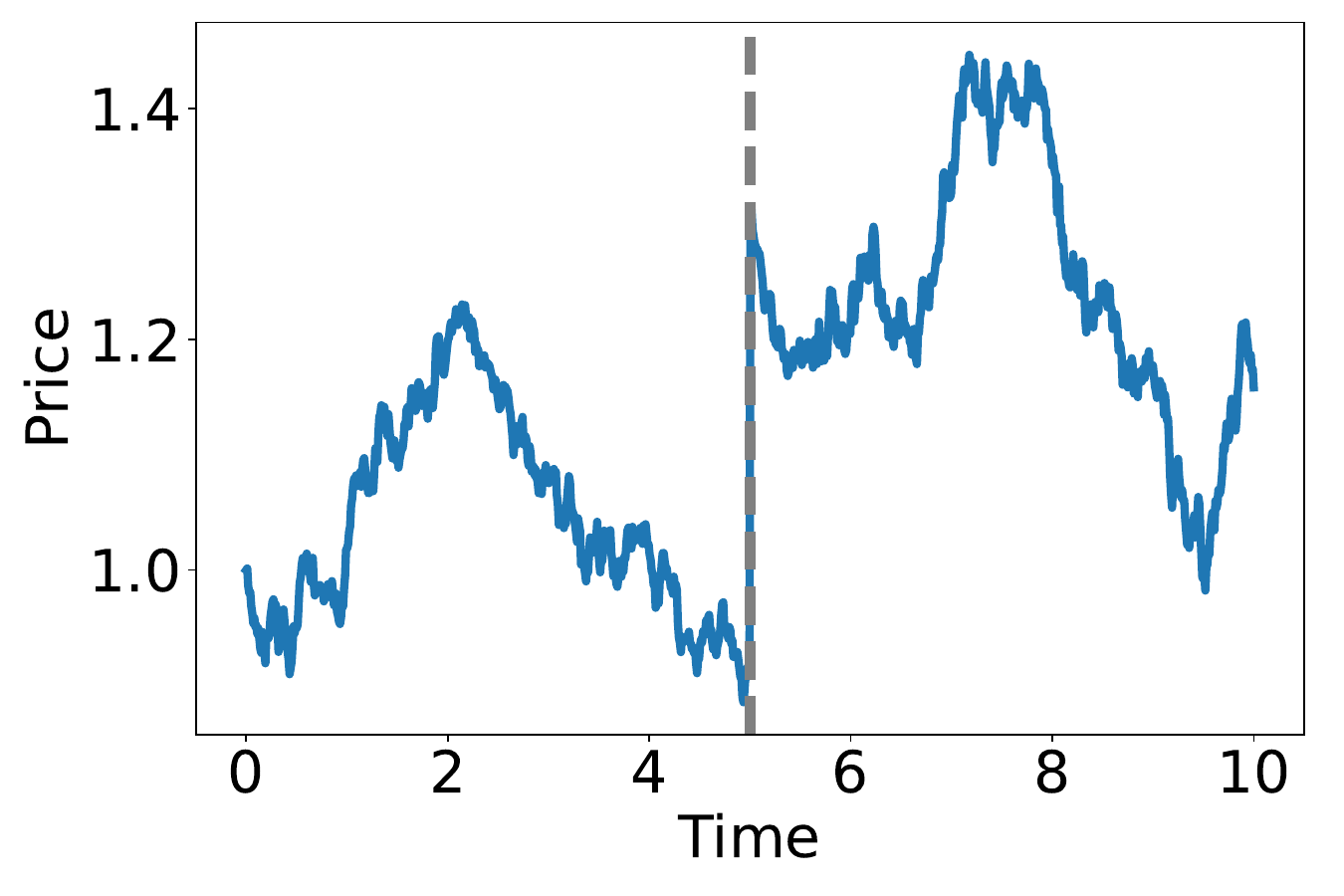}
    \caption{$\text{MJD}_{p_{no}}$}
  \end{subfigure}
  \begin{subfigure}[b]{0.24\linewidth}
    \includegraphics[width=\linewidth]{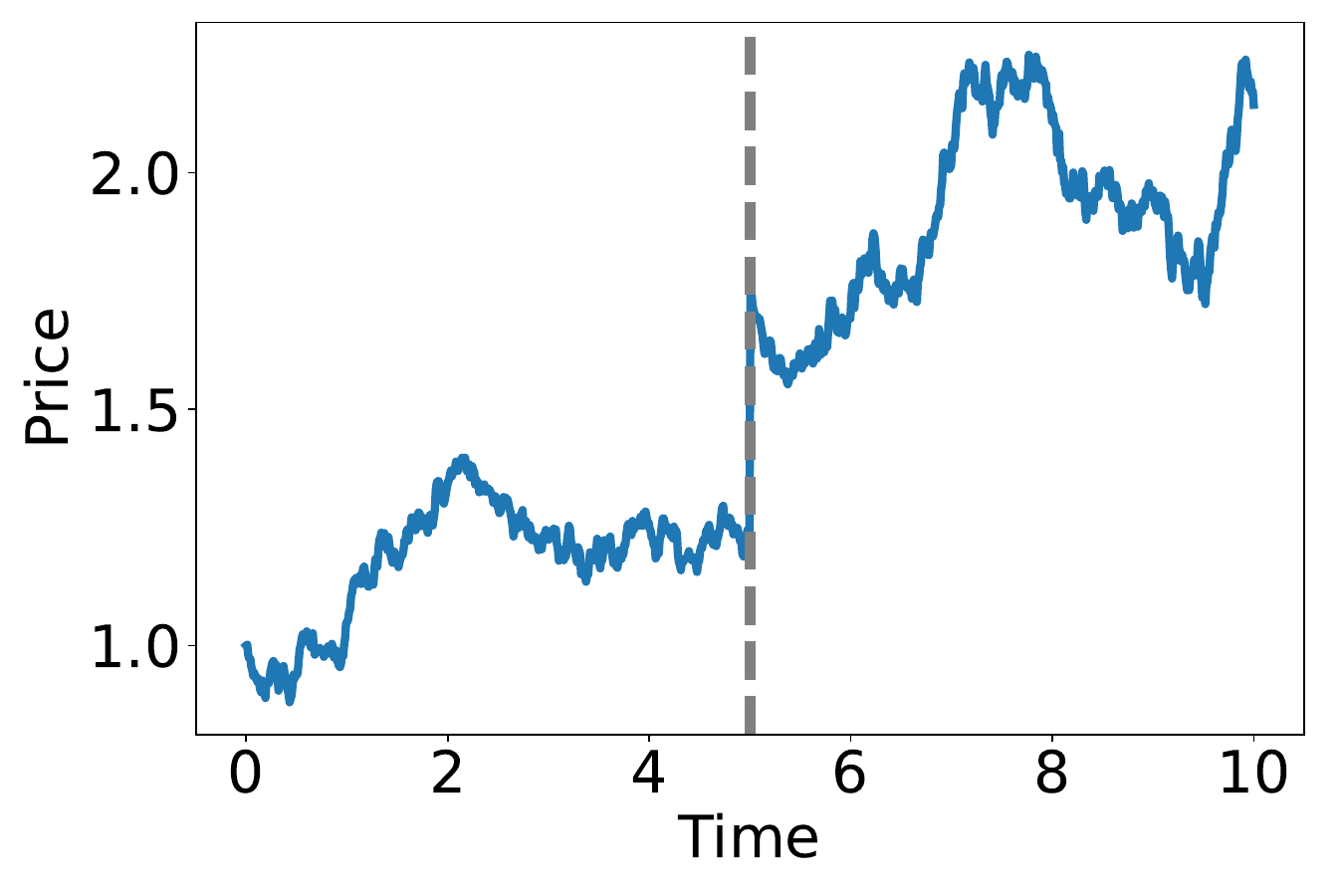}
    \caption{$\text{MJD}_{p_{up}}$ }
  \end{subfigure}
    \begin{subfigure}[b]{0.24\linewidth}
  \includegraphics[width=\linewidth]{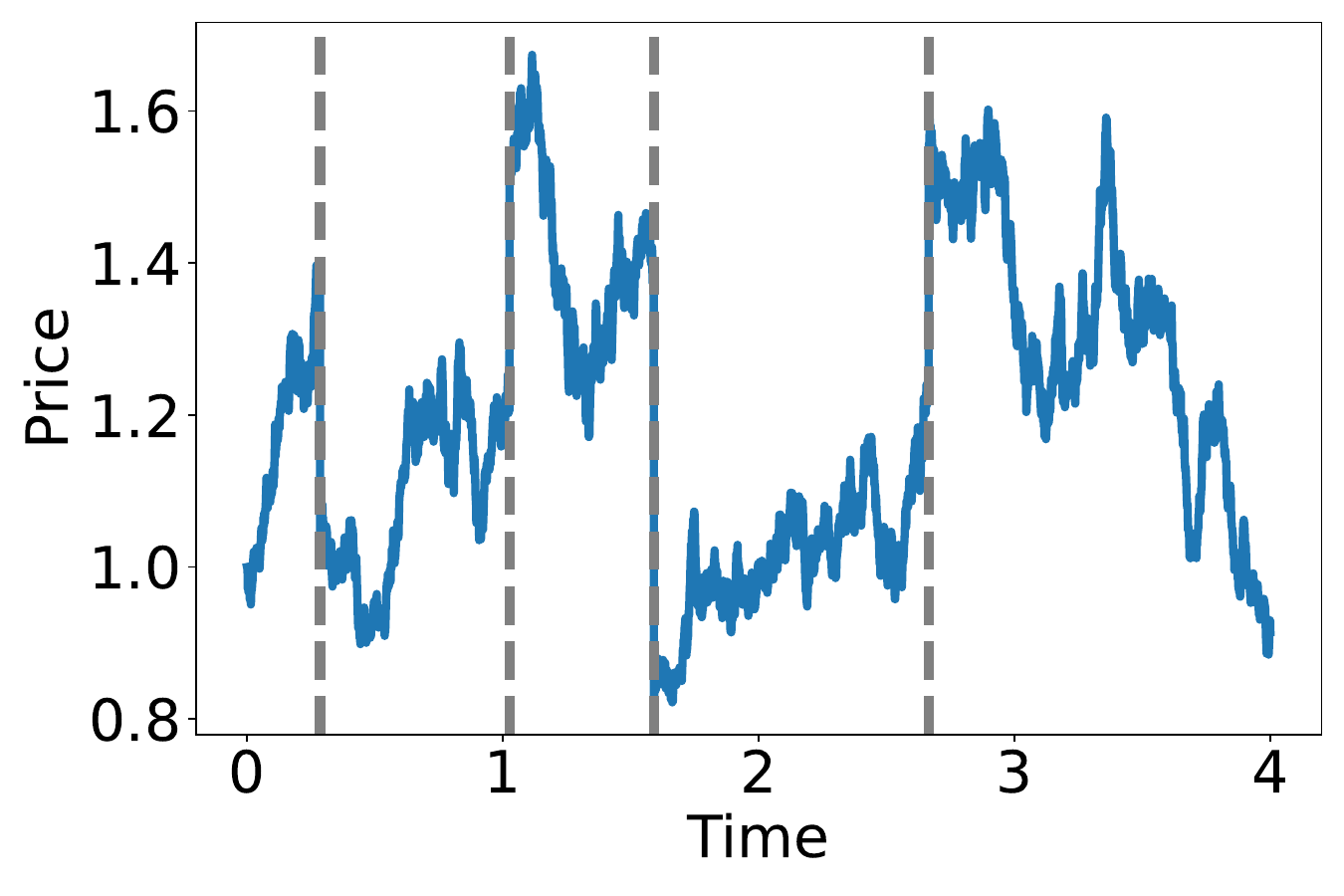}
    \caption{MCP }
  \end{subfigure}
  \caption{MJD Models. Blue lines depict MJD processes, and a grey dashed line marks the ground truth change-point $\tau = 5.7057$ for (a)-(d) and $\tau = 5.0$ for (e), (f). MCP (g) has 4 change-points, which are located at 0.2861, 1.0263, 1.5904, and 2.6647. Illustrated scenarios include (a) $\text{MJD}_{t_{no}}$: An MJD model without drift; (b) $\text{MJD}_{t_{up}}$: MJD model with a globally increasing trend; (c) $\text{MJD}_{t_{inv}}$:  MJD characterized by an increasing trend prior to the shift, followed by a diverging trend post-shift; (d) $\text{MJD}_{t_{down}}$: MJD with an upward trend but a downward shift; (e) $\text{MJD}_{p_{no}}$: MJD with polynomial drift; (f) $\text{MJD}_{p_{up}}$: MJD model with polynomial drift and a globally increasing trend; (g) MCP (K = 4).}
  \label{fig:JDP}
\end{figure}

\begin{table}[h]
\caption{Synthetic Functions Parameters for Different MJD Models.}
\label{MJD-para}
\centering

\subcaption*{Linear jump-diffusion models}
\begin{tabular}{l c c c c c c}
\hline
& {$\mu_{\text{pre}}$} & {$\mu_{\text{post}}$} & {$\sigma$} & {$\lambda$} & {$\alpha$} & {$\delta$}\\
\hline
$\text{MJD}_{t_{no}}$   & 0.00 & 0.00 & 0.10 & 0.18 & 0.10 & 0.10 \\
$\text{MJD}_{t_{up}}$   & 0.10 & 0.10 & 0.10 & 0.18 & 0.10 & 0.10 \\
$\text{MJD}_{t_{inv}}$  & 0.10 & -0.05 & 0.10 & 0.18 & 0.10 & 0.10 \\
$\text{MJD}_{t_{down}}$ & 0.06 & 0.06 & 0.10 & 0.18 & -0.10 & 0.01 \\
MCP                    & 0.30 & -0.40 & 0.30 & 0.90 & -0.10 & 0.20 \\
\hline
\end{tabular}

\bigskip
\subcaption*{Nonlinear jump-diffusion models}
\begin{tabular}{ c c c c}
\hline
 & $f(\cdot)$ & $g(\cdot)$ & $h(\cdot)$\\ 
\hline
$\text{MJD}_{p_{no}}$ & 0.01$S_{t}$(1-$S_{t}$) & 0.1$\sqrt{S_{t}}$ & 0.10\\
$\text{MJD}_{p_{up}}$ & 0.1$S_{t}$(1-$S_{t}$)+0.041t & 0.14$\sqrt{S_{t}}$ & 0.10\\
\hline
\end{tabular}

\end{table}

To evaluate our proposed algorithm, we utilize the Merton Jump-Diffusion (MJD) model, which was originally for modeling stock price $S_t$ at time $t$. Here, we simply use this flexible model to simulate different underlying process scenarios for evaluating the performance of GDCPD. The MJD model is governed by a Stochastic Differential Equation (SDE) as follows:
\begin{equation}
d S_t=\mu S_{t} d t+\sigma S_{t} d W_t+S_{t} d J_t.
\end{equation}
Here, $\mu$ is the diffusion drift, $\sigma$ represents the diffusion's volatility, $\{W_t\}_{t\geq0}$ is a standard Brownian motion, and $J_t= \sum_{i=1}^{N_t}Y_i$ is a compound Poisson process. The jump sizes $Y_i$ are identically and independently distributed with a distribution of $N(\alpha, \delta^2)$, and the number of jumps $N_t$ follows a Poisson process with jump intensity $\lambda$. The process $S_t$ follows a geometric Brownian motion between jumps. This SDE has an exact solution, which is $S_t=S_0 \exp \left\{\mu t+\sigma W_t-\sigma^2 t / 2+J_t\right\}$. For the nonlinear jump-diffusion model, we adopt:
\begin{equation}
d S_t=f(S_{t},t) d t+g(S_{t},t) d W_t+h(S_{t},t) d J_t,
\end{equation}
where $f(S_{t},t)$ is the drift term, $g(S_{t},t)$ represents the diffusion volatility, and $h(S_{t},t)$ represents the jump size.

\begin{figure}[t]
    \centering
    \begin{subfigure}[b]{0.49\linewidth}    
    \includegraphics[width=\linewidth]{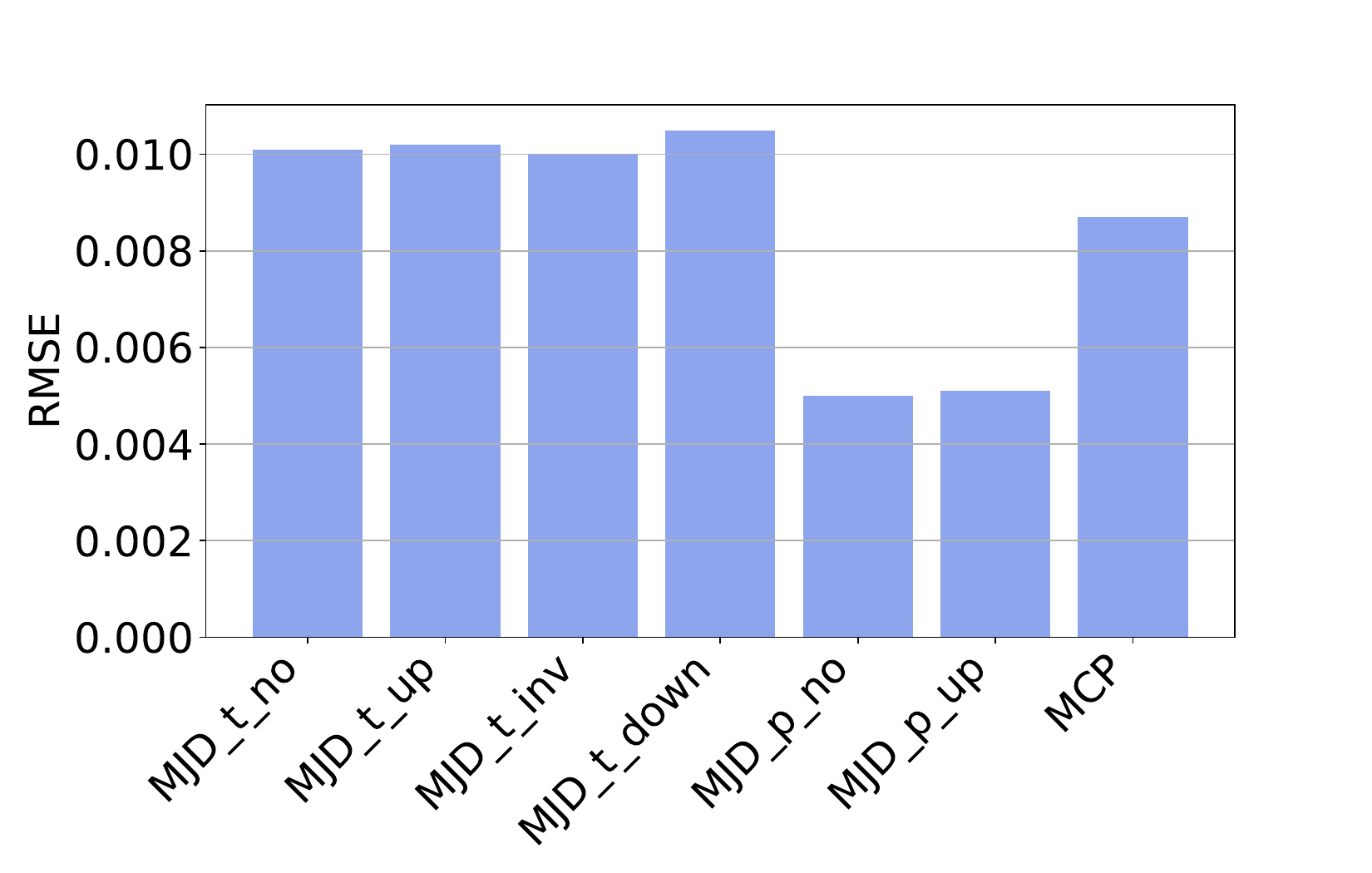}
  \subcaption{Simulation Experiments}
  \label{fig:C.10a}
  \end{subfigure}
    \begin{subfigure}[b]{0.49\linewidth}    
    \includegraphics[width=\linewidth]{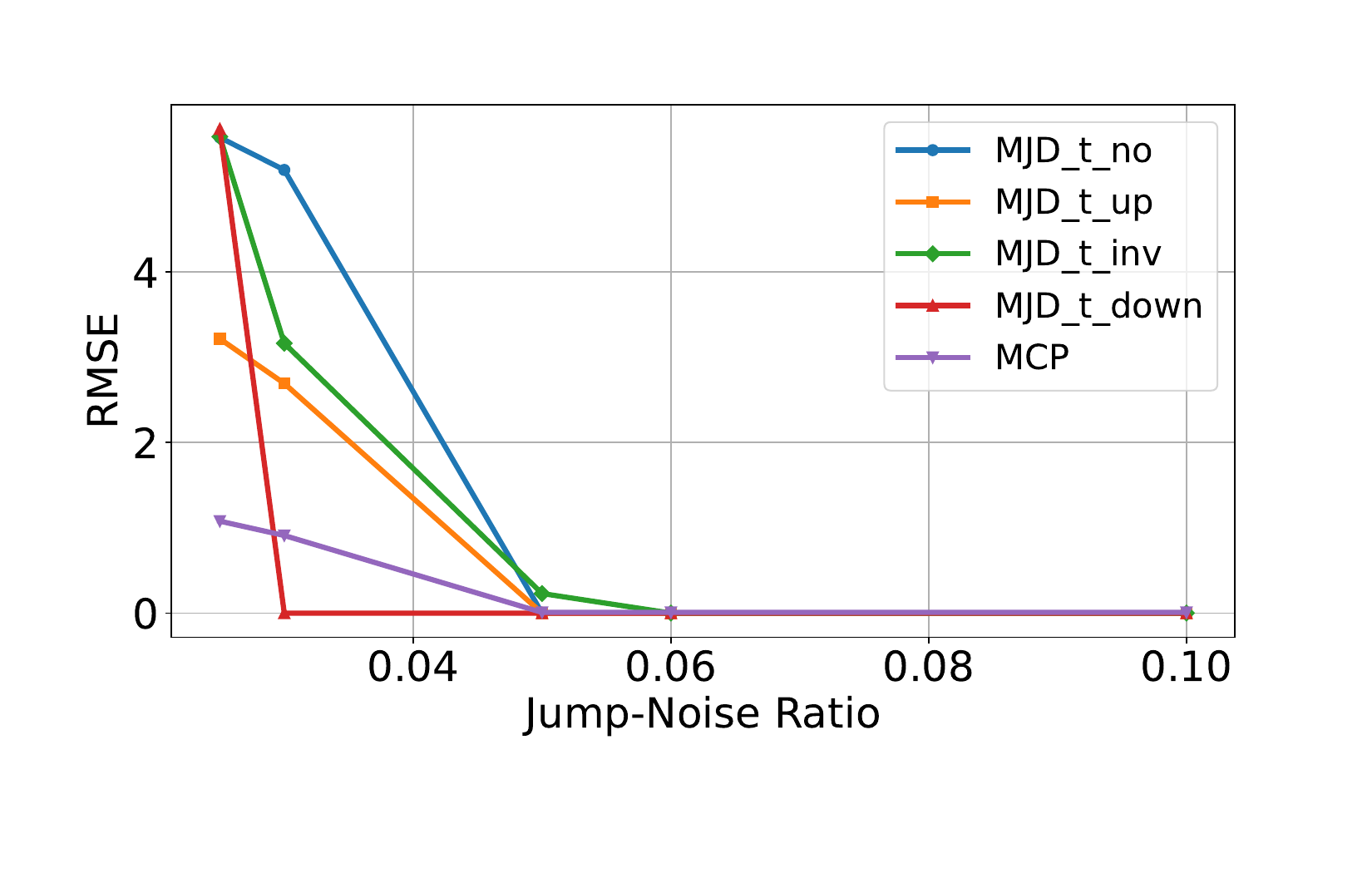}
  \subcaption{Jump-Noise Ratio Experiments }
  \label{fig:C.10b}
  \end{subfigure}
    \caption{RMSE Results for Simulation Experiments and Jump-Noise Ratio Experiments. (a): The RMSE of different MJD models. (b): The RMSE with different jump sizes and noise levels for MJD models.}
  \label{fig:SIMULATION}
\end{figure}

We assume the initial observation to be $S_0=1$ and set the time horizon $T=10$. By setting a time step of $dt = 1e-2$, we generate a total of 1000 simulated observations. We subsequently create seven distinct scenarios using the MJD model, each characterized by different patterns and abrupt change behaviors, as depicted in Figure~\ref{fig:JDP}: (1) $\text{MJD}_{t_{no}}$ features no trend; (2) $\text{MJD}_{t_{up}}$ displays an increasing trend; (3) $\text{MJD}_{t_{inv}}$ exhibits diverging trends before and after the change-point; and (4) $\text{MJD}_{t_{down}}$ undergoes a downward shift. Additionally, we examine two polynomial drift variants: (5) $\text{MJD}_{p_{no}}$, which integrates a polynomial drift function, and (6) $\text{MJD}_{p_{up}}$, which combines a polynomial drift with a globally increasing trend. And (7), a MCP case, where $T=4$ and time step $dt = 1e-3$. The specific model parameters employed for creating these simulation scenarios are detailed in Table~\ref{MJD-para}. And the RMSE results are shown in Figure~\ref{fig:C.10a}. The histogram indicates that the prediction errors for all models are generally low, with only minor variations in accuracy.

We performed a sensitivity analysis to evaluate the influence of the Jump-Noise Ratio (JNR), which is defined as:
\begin{equation}
    \text{JNR}_\text{t}=\cfrac{\lambda\alpha}{\sigma}.
\end{equation}
The JNR measures the relative contribution of jumps by comparing their average impact (considering both their frequency $\lambda$ and size $\alpha$) to the diffusion process volatility $\sigma$. A higher value of the JNR indicates that the jumps are easier to detect. From Figure~\ref{fig:C.10b}, the RMSE values are consistently low for all models when the JNR is below approximately 0.05.

\subsection{Parameter Choices}

\begin{table}[t]
\centering
\caption{Parameter Selection Results for Feature Relevance Threshold and Sliding Window Size}
\label{Table:Parameters}
\subcaption*{ARD length-scale $\lambda_d$ threshold for feature relevance}
\begin{tabular}{cccc}
    \hline
      $\lambda_d$ threshold  & 0.3 & 0.45 & 0.6\\
       \hline
        RMSE & 0.6326 & 0.1560  & 0.2082\\
        SF & 0.3077 & 0.1206 &  0.1986\\
        \hline
    \end{tabular}
\bigskip
\subcaption*{Sliding window size for online monitoring}
\begin{tabular}{ccccc}
    \hline
      Window Size A & 4 mins & 6 mins & 8 mins & 10 mins\\
       \hline
        RMSE & 0.2385 & 0.1560  & 0.3855 & 0.8416\\
        SF & 0.2027 & 0.1206 &  0.4206 & 0.6147 \\
        \hline
    \end{tabular}
\end{table}

We conducted experiments to justify the selection of key parameters for the algorithm. We focused on the ARD length-scale threshold $\lambda_d$ for feature relevance and the sliding window size in the online monitoring framework (Table~\ref{Table:Parameters}). For the ARD length-scale threshold, we tested values of 0.3, 0.45, and 0.6, and found that 0.45 provided the lowest RMSE and SF. Specifically, a threshold of 0.2 resulted in only 5 relevant features, which excluded important ones and increased RMSE, whereas a threshold of 0.6 included 50 relevant features, which introduced more noise and degraded performance. For the sliding window size, we evaluated 4, 6, 8, and 10 minutes and found that a 6-minute window minimized both RMSE and SF. Smaller windows led to overreaction to fluctuations, while larger windows smoothed out important variations and reduced accuracy. We validated that an ARD threshold of 0.45 and a window size of 6 minutes is the optimal combination for precise and interpretable model behavior in our algorithm setting.


\end{document}